\documentclass[oneside,final]{csri24}

\usepackage[top=1.355in,
			bottom=1.355in,
			left=1.5in,
			right=1.5in]{geometry}

\usepackage{amsfonts,
            amsmath,
            amsthm,
            graphicx,
            url,
            boldtensors}
\usepackage{subcaption}
\usepackage{xcolor}
\usepackage{float}

\theoremstyle{plain}
\newtheorem{remark}{Remark}

\title{Adaptive hybrid coupling with operator inference, the overlapping Schwarz alternating method and reinforcement learning}

\author{Trishit Mondal\thanks{Worcester Polytechnic Institute, tmondal@wpi.edu} \and Irina Tezaur\thanks{Sandia National Laboratories, ikalash@sandia.gov} \and Anthony Gruber\thanks{Sandia National Laboratories, adgrube@sandia.gov}}

\begin{document}

\maketitle

\begin{abstract}
Hybrid domain decomposition methods provide a flexible framework for coupling full order models (FOMs) and reduced order models (ROMs), but typically assume that the model assigned to each subdomain is selected \textit{a priori} and remains fixed throughout a simulation. This can be limiting for transient problems in which localized features propagate through the computational domain and the regions requiring high-fidelity resolution change over time. We introduce a reinforcement learning- (RL)-based approach for online adaptation of FOM-ROM models coupled using the overlapping Schwarz alternating method (O-SAM), an iterative domain decomposition method that solves subdomain-local problems while exchanging solution information through transmission boundary conditions on the overlapping subdomain interfaces. Deep Q-networks (DQNs) are trained offline to select among subdomain-local FOMs and pre-trained Operator Inference (OpInf) ROMs using a reward that balances solution accuracy and computational cost, while simultaneously penalizing unnecessary model switching. Once trained, the resulting policies are deployed predictively on problem instances not encountered during RL training without requiring a reference FOM solution.  We demonstrate the proposed approach on two numerical examples: a one-dimensional advection-diffusion problem with a moving front and a three-dimensional linear elastic wave propagation problem implemented in the {\tt Norma.jl} solid mechanics code.  For the advection-diffusion benchmark, the learned policy dynamically allocates high-fidelity resolution as the front propagates and provides a favorable accuracy-cost tradeoff relative to static FOM/ROM assignments; allowing the agent to additionally adapt the domain decomposition provides no further benefit. For the elastic wave benchmark, the learned policies for both  two and three subdomain decompositions track the propagating wave by assigning FOMs to the subdomains containing the wave and ROMs elsewhere, as expected. Importantly, this behavior is observed for problem instances not used during training. Our results demonstrate the potential of RL to enable predictive online adaptation of model fidelity within Schwarz-based hybrid simulations.

\end{abstract}

\section{Introduction} \label{sec:intro}

Multi-scale, multi-physics modeling and simulation (ModSim) is essential for the design and qualification of engineered components and systems relevant to Sandia’s mission. Analysts face significant delays due to both the mesh generation step of the ModSim workflow and the long run-time requirements for simulations, which can preclude multi-query analyses such as design optimization, control and  uncertainty quantification (UQ). Emerging approaches for building data-driven reduced order models (ROMs) have promised to reduce the runtime burden of ModSim and enable multi-query analyses. However, ROMs can suffer from their own deficiencies, including robustness, stability, and accuracy concerns, lengthy implementations, and a lack of systematic refinement mechanisms.

\begin{figure}[h]
    \centering
    \begin{subfigure}[t]{0.49\linewidth}
    \centering
    \includegraphics[width=\linewidth]{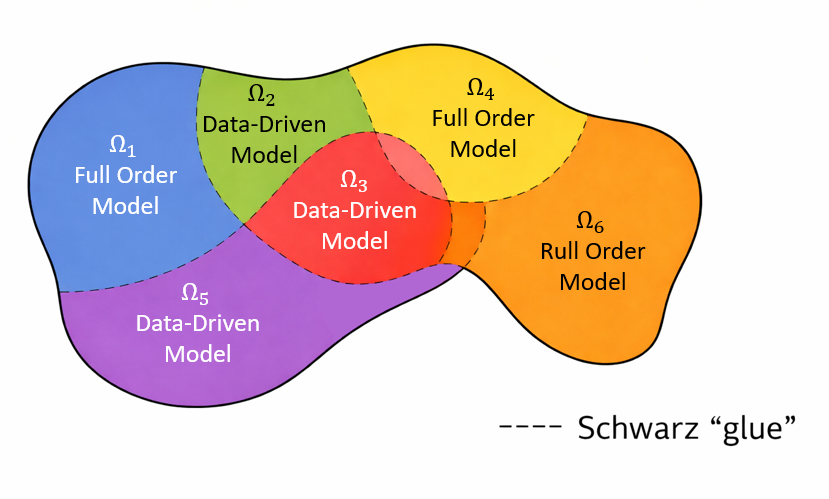}
    \caption{DD and model assignment}
  \end{subfigure}\hfill 
  \begin{subfigure}[t]{0.51\linewidth}
    \centering
    \includegraphics[width = \linewidth]{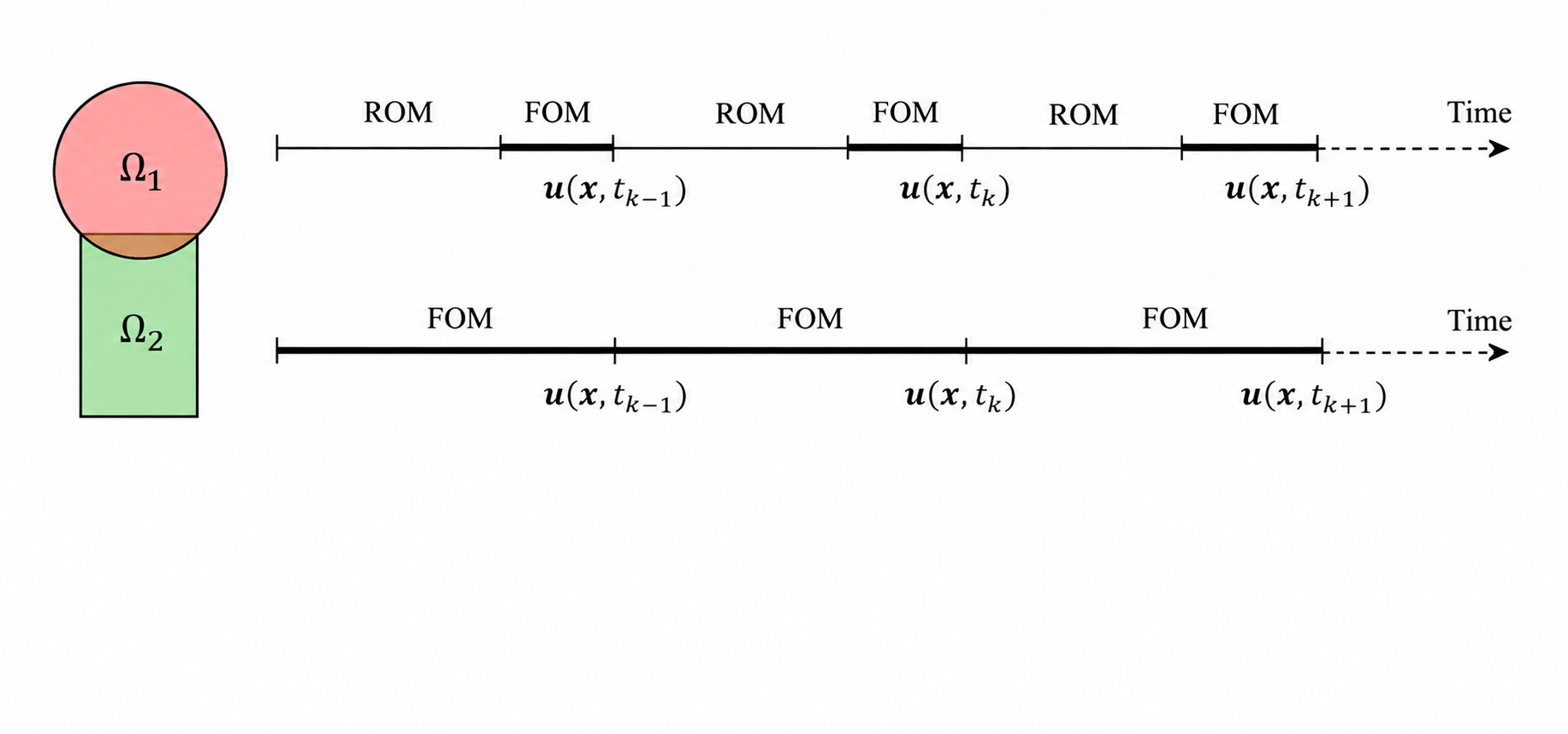}
    \caption{Online ROM-FOM-ROM switching}
  \end{subfigure} 
    \caption{Domain decomposition-based hybrid coupling and online ROM-FOM-ROM switching within a SAM coupling framework.
   }
    \label{fig:DD}
\end{figure}

In recent years, Sandia has expended a large effort to overcome both of the two aforementioned hurdles through the development of hybrid models based on optimal domain decomposition (DD) \cite{wentland2024Schwarz, Snyder:2023, Rodriguez:2025, tezaur2025hybrid}, as illustrated in Figure \ref{fig:DD}(a).  In our approach, the domain is decomposed into overlapping or non-overlapping subdomains according to some physics-based criteria, and different meshes and/or models are assigned to different subdomains according to these physics.  The subdomains are coupled via  an iterative technique known as the Schwarz alternating method (SAM) \cite{Schwarz:1870}.  The key idea behind SAM is to replace a global monolithic problem in a domain $\Omega$ by a sequence of smaller problems in subdomains $\Omega_i$, where $\Omega = \cup_{i}\Omega_i$.  The coupling happens through carefully defined transmission boundary conditions prescribed on the interfaces between the subdomains $\Omega_i$.  In our past work \cite{Mota:2017, Mota:2022, Snyder:2023, wentland2024Schwarz, Rodriguez:2025, tezaur2025hybrid}, we have shown that SAM is capable of seamlessly ``gluing together" arbitrary combinations of subdomain-local full order models (FOMs) and subdomain-local ROMs in a plug-and-play fashion.  Since the method does not require conformal meshing of the subdomains $\Omega_i$, it is extremely effective at simplifying meshing workflows.  The introduction of ROMs into the workflow has the potential to improve the predictive viability of the reduced models by enabling their spatial localization via DD, as well as through the online integration of high-fidelity information via FOM coupling.

Until now, our SAM-based coupling workflow has assumed that the DD and the assignment of meshes and/or models to individual subdomains are performed once, prior to the start of a simulation, and remain fixed throughout the coupled simulation. The primary limitation of this approach is that it cannot account for changing dynamics or traveling features that emerge and propagate during a simulation. For example, consider a traveling shock that requires a FOM to be accurately resolved. If the shock enters subdomains to which ROMs were assigned \textit{a priori}, inaccuracies may be introduced and subsequently propagate through the coupled system, degrading the solution over time.

The idea of dynamically switching between ROMs and FOMs online, as illustrated in Figure \ref{fig:DD}(b), is relatively new, but has been explored in several previous works. In \cite{Radermacher:2014, Corigliano:2015}, which consider solid mechanics problems, ROM-FOM switching is determined by the onset of plasticity within a prescribed subdomain: a ROM is employed as long as the subdomain remains in the elastic regime, and the subdomain switches to a FOM once plasticity is detected. Because plasticity is irreversible, switching is one-way, from ROM to FOM.

Several more recent works pursue closely related ideas, including adaptive enrichment, fidelity selection, and heterogeneous ROM-FOM coupling, without performing dynamic online ROM-FOM switching in the strict sense. In \cite{Smetana:2023}, Smetana and Taddei propose an adaptive localized ROM in which residual-based error indicators identify components requiring enrichment and high-fidelity local correction problems are solved to augment the corresponding reduced spaces. Thus, rather than switching a component from a ROM to a FOM, their approach selectively introduces high-fidelity computations online when the ROM is deemed insufficient. In \cite{Huang:2022}, Huang and coauthors develop adaptive and component-based ROM frameworks that incorporate online basis adaptation and permit heterogeneous coupling of ROM and FOM components; however, components are not dynamically switched between ROM and FOM representations during the simulation. In \cite{Ebrahimi:2024}, Ebrahimi and Yano develop a component-based hyper-reduced ROM that adaptively selects the hyper-reduction fidelity of individual components online based on a system-level error estimate. This constitutes online adaptive fidelity selection rather than ROM-FOM switching, as the components remain reduced order throughout the computation.
Finally, Hedayat et al. \cite{Hedayat2026AdaptiveNIROM} perform online temporal switching between ROM and FOM computations, periodically invoking the FOM to acquire high-fidelity data used to adapt the ROM before returning to reduced order prediction. In contrast to the present setting, their approach considers a single computational domain and therefore does not involve domain decomposition or coupling between distinct ROM and FOM subdomains.

One fundamental challenge in enabling online ROM-FOM switching is determining when and where a change in model fidelity should occur. Fundamentally, this problem is related to extrapolation detection in machine learning and requires some measure of model adequacy.  While there has been some work on error estimation for certain types of ROMs, e.g., projection-based ROMs \cite{BLONIGAN2023115988}, reliable error indicators are not readily available for many classes of data-driven models. Herein, we address this challenge by leveraging reinforcement learning (RL) \cite{sutton2018reinforcement} to learn an online model-switching policy for subdomain-local FOMs and Operator Inference (OpInf) ROMs \cite{willcox2016opinf} coupled via overlapping SAM (O-SAM). 
We refer to the resulting RL-based adaptive O-SAM framework as ``RL--O-SAM".  
In our workflow, the RL agent is trained on a collection of problem instances to select the model employed in each subdomain over time, with a reward function that balances computational cost, solution error, and penalizes unnecessary model switching. Once trained, the learned policy is deployed predictively on problem instances not encountered during RL training, including problems characterized by different parameters, boundary conditions, and/or initial conditions from those used to train the policy. In this way, the proposed framework seeks to learn when and where high-fidelity resolution is required while exploiting computationally inexpensive ROMs elsewhere, enabling the model assignment within the domain decomposition to adapt dynamically as the solution evolves.

Toward this effect, the remainder of this paper is organized as follows.  Section \ref{sec:opinf-osam} provides some preliminaries to keep this paper self-contained, overviewing the OpInf approach to model order reduction and the O-SAM algorithm applied to couple subdomain local OpInf ROMs with each other or with subdomain-local FOMs.  Section \ref{sec:adapt-rl} overviews RL and describes how it is utilized to perform online adaptation of O-SAM.  Section \ref{sec:results} presents some numerical results illustrating the accuracy and efficiency of the proposed approach, applied to an advection-diffusion benchmark with a moving shock, and to a linear elastic wave propagation problem in solid mechanics.  Finally, conclusions are offered in Section \ref{sec:conc}. 

\vspace{0.5em}

\begin{remark}
We note that the ROMs being coupled in our approach are all pre-trained and not updated on-the-fly using newly-available FOM information.  Performing online OpInf ROM updates, a relatively open topic, will be the subject of future work, and may leverage techniques from the recent pre-print by Koike et al. \cite{koike2026streamingoperatorinferencemodel}, in which a streaming OpInf approach is developed. 
\end{remark}

\section{Hybrid coupling with operator inference (OpInf) and the overlapping Schwarz alternating method (O-SAM)} \label{sec:opinf-osam}

As discussed above, this paper presents an RL-based approach for performing online switching between subdomain-local FOMs and OpInf ROMs \cite{willcox2016opinf}, coupled via O-SAM \cite{Schwarz:1870}.  Before describing our RL-based workflow (Section \ref{sec:rl-osam}), we briefly describe OpInf model order reduction and its combination with O-SAM \cite{tezaur2025hybrid}.

\subsection{Operator Inference (OpInf)}  First proposed by Peherstorfer and Willcox in \cite{willcox2016opinf}, OpInf is a projection-inspired model reduction approach that is fully non-intrusive, requiring no access to the underlying high-fidelity simulation code.  The approach consists of three steps: (i) the creation of a reduced basis ${\bf \Phi}_r$  from snapshot data collected during a high-fidelity simulation, (ii) the solution of a regularized least-squares optimization problem for the operators defining the ROM, and (iii) online deployment of the learned ROM.  

Following the approach in \cite{tezaur2025hybrid}, we construct our reduced basis ${\bf \Phi}_r$ using a common technique known as Proper Orthogonal Decomposition (POD) \cite{Holmes:1996}.  Assume, without loss of generality, that we have collected $K+ 1$ snapshots of a solution $~u(t)$ to a given transient partial differential equation (PDE) at times $T_0, T_1, ..., T_K$ and placed them in the columns of a snapshot matrix $~U$, i.e., 
\begin{equation} \label{eq:snap}
    ~U:=\left( \begin{array}{cccc} ~u(T_0), & ~u(T_1) &, \cdots, & ~u({T_K}) \end{array} \right) \in \mathbb{R}^{N\times K},
\end{equation}
where $N$ is the size of each snapshot, corresponding to the number of degrees of freedom (dofs) in the PDE being solved.
POD is based on the assumption that the solution to a given PDE has a low rank relative to the FOM dof space $\mathbb{R}^N$, so that the snapshot matrix $~U$
admits a low-rank approximation. POD begins by performing a thin 
singular value decomposition (SVD) of $~U$, so that $~U = ~\Phi ~\Sigma ~\Psi^T$.  
The first $r \leq \operatorname{rank}(~U)$ left singular vectors form the 
rank-$r$ POD basis $~\Phi_r \in \mathbb{R}^{N\times r}$. This basis provides 
the optimal rank-$r$ approximation of the snapshots in the Frobenius norm, so that 
\begin{equation}
     \|~U-~\Phi_r~\Phi_r^\top ~U\|_F^2
    = \sum_{i=r+1}^{\operatorname{rank}(~U)} \sigma_i^2,
\end{equation}
where $\sigma_i$ denotes the $i$th singular value of $~U$.  The reduced
dimension $r$ is commonly selected to retain a prescribed fraction
$\xi \in (0,1)$ of the snapshot energy, i.e., as the smallest integer
satisfying
\begin{equation}
    \frac{\sum_{i=1}^{r}\sigma_i^2}
         {\sum_{i=1}^{\operatorname{rank}(~U)}\sigma_i^2}
    \geq \xi.
\end{equation}

OpInf is founded on the observation that projection-based reduction of a polynomial FOM preserves its polynomial structure.  For example, consider a linear FOM of the form 
\begin{equation} \label{eq:fom}
    \dot{~u} + ~A ~u = ~B~g
\end{equation}
with $~u \in \mathbb{R}^N$, $~A \in \mathbb{R}^{N \times N}$, $~B \in \mathbb{R}^{N \times n}$ and $~g \in \mathbb{R}^n$,  for $N, n \in \mathbb{N}$.  A system of the form \eqref{eq:fom} is obtained by discretizing a linear PDE in space using the finite element method and enforcing Dirichlet boundary conditions using the $~B~g$ term, where the $~g$ vector encodes the Dirichlet boundary data\footnote{For a derivation  of how the $~B~g$ and $\bar{~B}~g$ terms encode Dirichlet boundary condition in both the ROM and the FOM, the reader is referred to \cite{tezaur2025hybrid}.}.
Projecting \eqref{eq:fom}
onto a reduced basis $~\Phi_r$ yields the linear ROM 
\begin{equation} \label{eq:rom}
     \dot{\widehat{~u}} + \widehat{~A} \widehat{~u} = \bar{~B}~g,
\end{equation}
where $\widehat{~u}:= ~\Phi_r^T ~u$, $\widehat{~A}:= ~\Phi_r^T ~A ~\Phi_r$ and $\bar{~B}: = ~\Phi_r^T ~B$. In traditional (intrusive) projection-based model order reduction, computing the $\widehat{~A}$ and $\bar{~B}$ operators in  \eqref{eq:rom} requires access to the matrix $~A$ and the term $~B~g$, and hence to the code that was used to generate these matrices.  Non-intrusive OpInf avoids this by learning the reduced operators in \eqref{eq:rom} directly from the snapshot data \eqref{eq:snap} by solving a convex least-squares minimization problem.
Given data for the projected state $\widehat{~U} = ~\Phi_r^T ~U$ and approximate derivative information, $D_t(\widehat{~U})$ 
computed with, e.g., a finite
difference operator $D_t$, the OpInf learning problem takes the form: 
\begin{equation}
    \underset{\widehat{~A},\bar{~B}}{\operatorname{arg\,min}}
    \left\|
        D_t(\widehat{~U})
        - \widehat{~A}\widehat{~U}
        - \bar{~B}~G
    \right\|_F^2
    +
    \eta
    \left(
        \|\hat{~A}\|_F^2
        + \|\bar{~B}\|_F^2
    \right).
    \label{eq:opinf-learning}
\end{equation}
Here, $~G$ is a matrix whose $i$th
column is $~g(T_i)$, and $\eta \in \mathbb{R}^{+}$ is a
(scalar-valued) regularization parameter controlling the condition number of the resulting linear system. 

\subsection{O-SAM applied to Operator Inference} \label{sec:o-sam}

\begin{figure}[htb!]
\centering
\includegraphics[width=\textwidth]{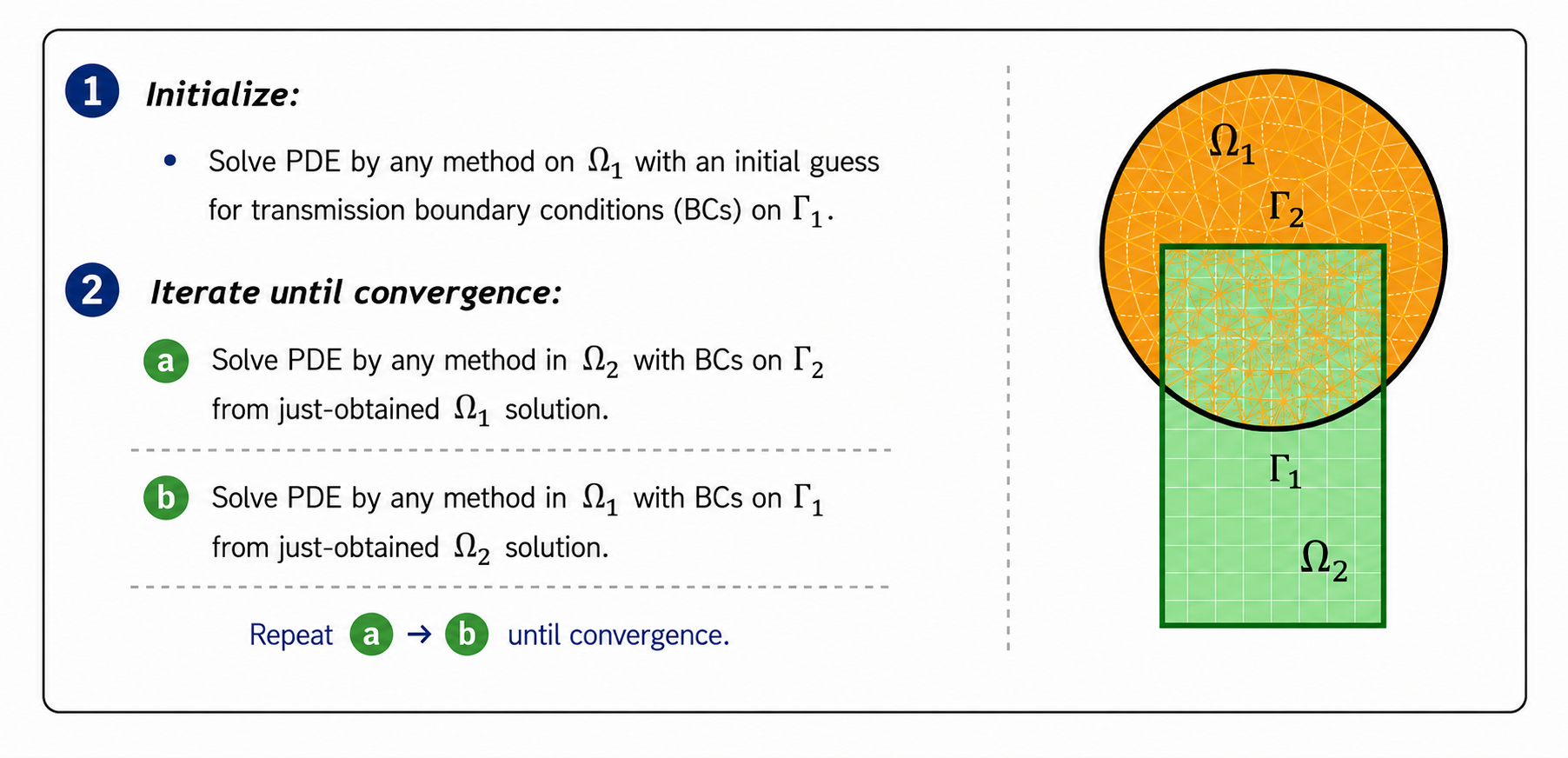}
\caption{The basic O-SAM algorithm for the coupling of two overlapping subdomains, $\Omega_1$ and $\Omega_2$.}
\label{fig:o-sam}
\end{figure}

The second ingredient in our workflow is O-SAM, which we use as a means to perform concurrent subdomain coupling.  Consider a physical domain $\Omega \in \mathbb{R}^d$, for $d=1,2,3$, and assume $\Omega$ has been 
decomposed into two overlapping subdomains, $\Omega_1$ and $\Omega_2$, such that $\Omega = \Omega_1 \cup \Omega_2$ and $\Omega_1 \cap \Omega_2 = \emptyset$, as shown in Figure \ref{fig:o-sam}.  Let the interior interface boundaries be defined as $\Gamma_1 := \partial \Omega_1 \cap \Omega_2$ and $\Gamma_2:=\partial \Omega_2 \cap \Omega_1$.  In the multiplicative O-SAM algorithm, illustrated pictorally in Figure \ref{fig:o-sam}, the governing PDE is solved by iterating sequentially between subdomain-local problems, with boundary information exchanged to ensure compatibility across the ``Schwarz boundaries" $\Gamma_1$ and $\Gamma_2$, as shown in Figure \ref{fig:o-sam}.  It has been shown \cite{Lions:1988, Mota:2017} that, for an overlapping DD, the Schwarz iteration process is guaranteed to converge if Dirichlet transmission boundary conditions (BCs) are specified on the Schwarz boundaries, provided the underlying problem is well-posed and the overlap region is non-empty.  

\begin{figure}[htb!]
\centering
\includegraphics[width=0.85\textwidth]{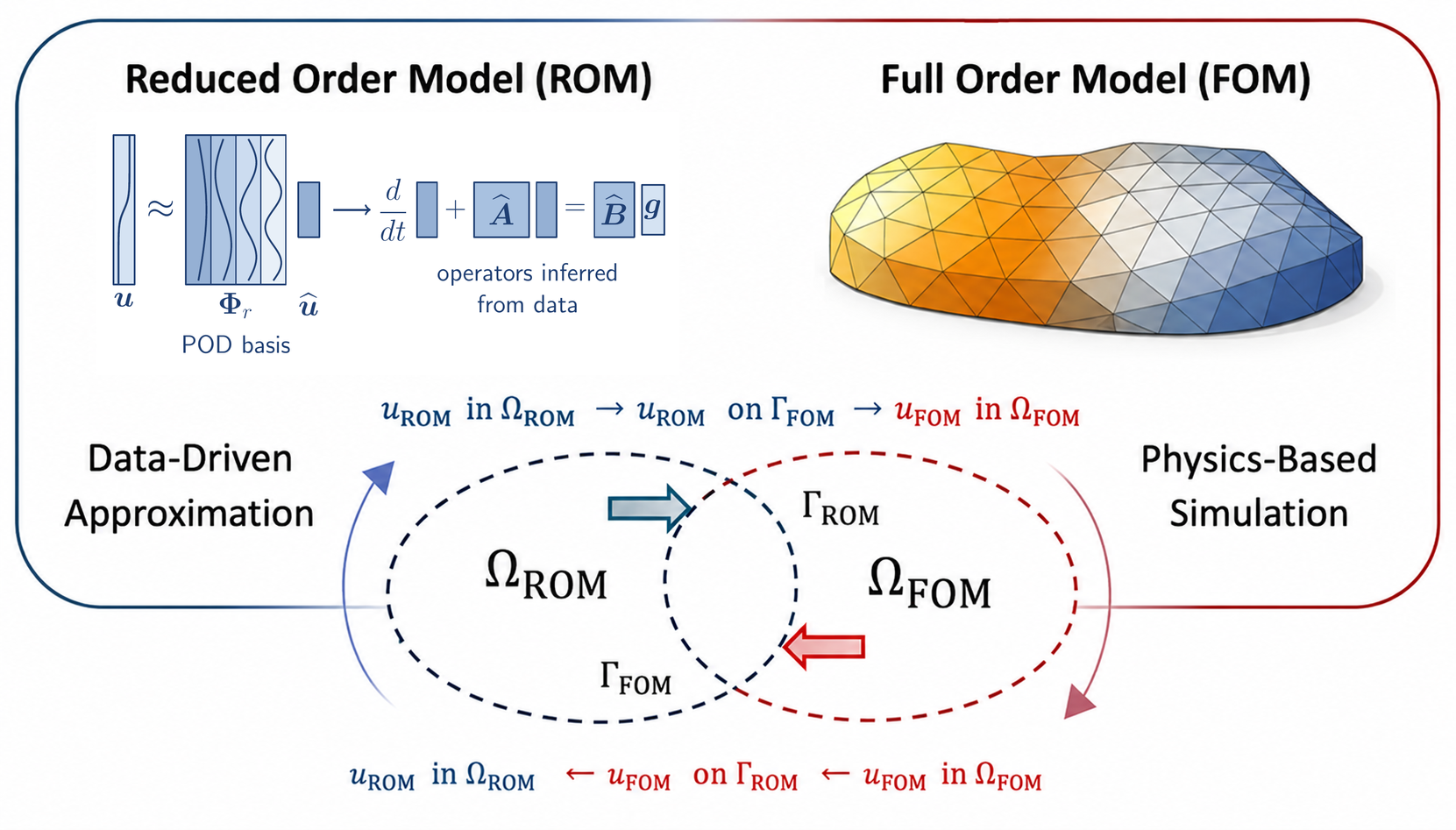}
\caption{Graphical depiction of O-SAM being used to couple a subdomain-local FOM with a subdomain-local OpInf ROM.}
\label{fig:o-sam_rom-fom}
\end{figure}

There is a critical detail implied by Figures \ref{fig:o-sam} and \ref{fig:o-sam_rom-fom}: the PDEs in each of the subdomains can be solved by \textit{any} method on \textit{any} mesh, meaning that the Schwarz algorithm allows for the plug-and-play coupling of a variety of models and meshes, including a combination of conventional and data-driven models.   Indeed, it has been shown in previous work that O-SAM is capable of coupling regions with
different mesh resolutions, different element types, different time integration schemes (e.g., implicit and
explicit), and even different models (e.g., FOM and ROM), all without introducing any artifacts exhibited
by alternative coupling methods \cite{Mota:2017, Mota:2022, Snyder:2023, wentland2024Schwarz, Rodriguez:2025, tezaur2025hybrid}.  In addition, the method is minimally intrusive to implement in existing HPC software frameworks and possesses rigorous convergence properties/guarantees \cite{Mota:2017, Mota:2022} in the all-FOM coupling setting. 

Herein, we adopt the O-SAM OpInf-OpInf and OpInf-FOM coupling methodology developed in \cite{tezaur2025hybrid} and sketched succinctly below.  The approach is depicted graphically in Figure \ref{fig:o-sam_rom-fom}.  
Consider the two subdomain decomposition shown in Figure \ref{fig:o-sam} and assume that an OpInf ROM is employed in $\Omega_1$ while a FOM is used in $\Omega_2$.  Suppose, additionally, that we have pre-learned the reduced basis and operators defining the former model by solving the minimization problem \eqref{eq:opinf-learning} restricted to $\Omega_1$.
For simplicity, we assume that there are no prescribed
essential boundary conditions on the physical boundary, so that the
only Dirichlet data imposed on each subdomain are the Schwarz
transmission conditions.
At the semi-discrete level, the corresponding subdomain models take the form: 
\begin{equation}
    \dot{\widehat{~u}}_1
    + \widehat{~A}_1 \widehat{~u}_1
    = \bar{~B}_1 ~g_1,
    \qquad
    \dot{~u}_2
    + ~A_2 ~u_2
    = ~B_2 ~g_2,
\end{equation}
where $~g_i$ contains the Dirichlet data imposed on the Schwarz boundary 
$\Gamma_i$. More precisely, let
\begin{equation}
    \mathcal{P}_{\Omega_j \to \Gamma_i} :
    \mathbb{R}^{N_j} \rightarrow \mathbb{R}^{n_{\Gamma_i}},
    \qquad i \neq j,
\end{equation}
denote an operator that restricts (or traces) the solution on $\Omega_j$ to the
Schwarz boundary $\Gamma_i$ of the neighboring subdomain, where $N_j \in \mathbb{N}^+$ is the number of dofs in $\Omega_j$ and $n_{\Gamma_i} \in \mathbb{N}^+$ is the number of dofs on $\Gamma_i$. Now, the
Dirichlet data imposed on $\Gamma_i$ are given by
\[
    ~g_i
    =
    \mathcal{P}_{\Omega_j \to \Gamma_i}~u_j,
    \qquad i \neq j.
\]
\begin{remark}
The operator $\mathcal{P}_{\Omega_j \to \Gamma_i}$ may additionally
incorporate interpolation when the discretizations of $\Omega_i$ and
$\Omega_j$ are nonconformal. We do not consider this case herein but it is considered in 
some of our past works on Schwarz coupling, e.g., \cite{Mota:2017, Mota:2022, wentland2024Schwarz, Mota2025Contact, tezaur2025hybrid}.  \\
\end{remark}

Suppose now that
$\boldsymbol{\Phi}_{r,1} \in \mathbb{R}^{N_1 \times r_1}$ is the size $r_1 \in \mathbb{N}^+$ POD basis
used to construct the OpInf model on $\Omega_1$, so that
\begin{equation}
      ~u_1(t)
    \approx
    \boldsymbol{\Phi}_{r,1}\widehat{~u}_1(t).
\end{equation}
Let $n \in \mathbb{N}$ denote the Schwarz iteration index. At Schwarz iteration $n$,
the coupled OpInf--FOM subdomain problems are given by
\begin{align} \label{eq:schwarz_iter}
    \dot{\widehat{~u}}_1^{(n)}
    + \widehat{~A}_1\widehat{~u}_1^{(n)}
    &=
    \bar{~B}_1
    \mathcal{P}_{\Omega_2 \to \Gamma_1}
    ~u_2^{(n-1)},
    \\
    \dot{~u}_2^{(n)}
    + ~A_2~u_2^{(n)}
    &=
    ~B_2
    \mathcal{P}_{\Omega_1 \to \Gamma_2}
    \boldsymbol{\Phi}_{r,1}\widehat{~u}_1^{(n)}.
\end{align}
for $n=0, 1, ...$.  The Schwarz iteration \eqref{eq:schwarz_iter} continues until the differences in the solutions $u_i^{(k+1)}$ and $u_i^{(k)}$ are sufficiently small, for $i =1, ..., n_d$, with $n_d \in \mathbb{N}^+$ denoting the number of subdomains (in this case, $n_d=2$), 
or the number of Schwarz iterations performed hits a pre-specified maximum Schwarz iteration value, {\tt maxit}$\in \mathbb{N}^+$.  
Following the application of a time-discretization scheme, at each time step, the subdomain-local OpInf and FOM problems are solved alternately until the Schwarz convergence criterion is satisfied. The converged solution is then advanced to the next time step, and this procedure is repeated until the final simulation time is reached.

The above workflow describes the online Schwarz coupling phase of the O-SAM algorithm, without considering the offline construction phase of the OpInf ROM in $\Omega_1$.  We construct our subdomain-local ROMs following the methodology described in Algorithm 2 in \cite{tezaur2025hybrid}, which is based on a top-down training approach.  In this approach: (i) a set of O-SAM-based FOM-FOM coupled simulations on $\Omega_1$ and $\Omega_2$ are performed, (ii) snapshots for $~U_1$ an $~G_1$, the solution and boundary data from $\Omega_1$ are saved, (iii) a POD basis, ${\bf \Phi}_{r,1}$ is computed, and (iv) the regularized minimization problem in \eqref{eq:opinf-learning}, restricted to $\Omega_1$, is solved to obtain the $\widehat{~A}_1$ and $\bar{~B}_1$ operators in \eqref{eq:schwarz_iter}.  

Extending the above algorithm to the case of OpInf-OpInf coupling and multiple subdomains is straightforward.  We note that various enhancements, such as the incorporation of essential boundary conditions on the outer boundaries of $\Omega_1$ and $\Omega_2$, 
and the reduction of the boundary dofs is also straightforward, but we omit these details herein for the sake of brevity.  The interested reader is referred to \cite{tezaur2025hybrid} for those and other details.

\section{Reinforcement learning for online adaptation of O-SAM}
\label{sec:adapt-rl}

The objective of the present work is to use RL  to enable online adaptation of the O-SAM coupling framework described in Section \ref{sec:o-sam}. 
In particular, we seek to learn a policy that dynamically selects between ROM and FOM representations within each subdomain as the solution evolves in time, thereby enabling both ROM-to-FOM and FOM-to-ROM switching during a simulation. We additionally consider extending this framework to allow the underlying DD itself to vary in time, so that both the subdomain partition and the model assigned to each subdomain may be selected adaptively. 

The remainder of this section introduces the RL methodology employed to learn these adaptive policies and describes its integration within the O-SAM iteration process.  In particular, Section \ref{sec:rl} provides an overview of reinforcement learning in a general setting, Section \ref{sec:dqn} describes the specific RL algorithm used in this work, namely the deep Q-network (DQN) approach, and Section \ref{sec:rl-osam} describes the application of DQN to the online model-switching and DD adaptation problems considered in this work.

\subsection{Reinforcement learning} \label{sec:rl}

RL \cite{sutton2018reinforcement, puterman2014markov} is a machine learning paradigm in which an agent learns to make sequential decisions by interacting with an environment, in our case, a numerical simulation. 
The RL decision-making process begins with the agent observing the current state $s_t\in\mathcal S$. Based on this state, the agent selects an action $a_t\in\mathcal A$ according to a policy $\pi(a\mid s)$ and receives a scalar reward $r_t\in\mathbb R$ quantifying the immediate benefit of that decision. The state represents the information available to the agent at the time a decision is made, while the action represents one of the available decisions. In the hybrid simulations considered herein, the state is constructed from the numerical solution at a given time, the actions correspond to different model selections, and the reward balances solution accuracy and computational cost, as detailed in Sections \ref{sec:rl-osam} and \ref{sec:results}.

Given a state $s_t$ and an action $a_t$, the environment  transitions to a new state $s_{t+1}$ according to the transition dynamics
\begin{equation} \label{eq:stplus1}
    s_{t+1} \sim \mathbb{P}(\cdot \mid s_t, a_t),
\end{equation}
where $\mathbb{P}$ denotes the state transition probability distribution.  It is evident from \eqref{eq:stplus1} that an action affects not only the reward received at the current step, but also the state from which subsequent decisions are made. This sequential dependence is central to RL: an action that appears favorable immediately may lead to less favorable outcomes later, and vice versa.  This sequential interaction between the agent and the environment is commonly modeled as a Markov Decision Process (MDP) \cite{puterman2014markov}, defined by the tuple
\begin{equation}
    (\mathcal{S}, \mathcal{A}, P, R, \gamma),
\end{equation}
where $\mathcal{S}$ is the state space, $\mathcal{A}$ is the action space, $P$ is the transition model, $\gamma \in [0,1)$ is the discount factor, and $R:\mathcal{S}\times\mathcal{A}\times\mathcal{S}\rightarrow\mathbb{R}$ is the \emph{reward function} generating the scalar reward $r_t=R(s_t,a_t,s_{t+1})$ received at every step. 


The objective of an RL agent is to learn a policy that maximizes the expected cumulative discounted reward,
\begin{equation} \label{eq:J}
    J(\pi) = \mathbb{E}_{\pi}\left[\sum_{t=0}^{\infty}\gamma^t r_t\right],
\end{equation}
where the expectation is taken over trajectories induced by the policy, the transition dynamics, and the initial state distribution. The discount factor $\gamma$ determines the relative weighting of immediate and future rewards, so that values of $\gamma$ close to zero place greater emphasis on immediate rewards, whereas values closer to one place greater weight on the long-term consequences of the agent's decisions.
In \eqref{eq:J}, $J(\pi)$ should be distinguished from the reward function $R$: whereas $R$ defines the reward received at each step, $J(\pi)$ measures the expected cumulative performance of a policy $\pi$.


In modern RL algorithms, neural networks are often used to approximate the functions that define the RL agent's decision-making policy, providing a practical means of representing these functions over high-dimensional state spaces. In the present work, we employ such a neural-network-based RL approach to enable adaptive decision-making within an evolving numerical simulation. Specifically, we augment the O-SAM coupling algorithm described in Section \ref{sec:opinf-osam} with an RL-based model-switching strategy that adaptively selects between FOM and ROM subdomain models as the simulation evolves.  The details of this approach are given in Section \ref{sec:rl-osam}, following the description  of the specific value-based RL algorithm employed in this work, the deep Q-network (DQN) \cite{mnih2013playing} (Section \ref{sec:dqn}).


\subsection{Deep Q-networks (DQNs)} \label{sec:dqn}

As stated above, the RL policy learned herein utilizes a DQN \cite{mnih2013playing}.  DQNs are particularly well-suited to problems with high-dimensional state spaces and a discrete set of available actions. This makes DQN a natural choice for the model selection problem considered here, in which the state contains high-dimensional information describing the evolving solution, while the available actions correspond to a discrete and relatively small set of model assignments.\\

\begin{remark}

The reader may wonder why we do not consider simpler RL approaches that are not based on neural networks, such as tabular Q-learning \cite{sutton2018reinforcement}. Tabular
methods maintain a specific value estimate for every state-action pair and are effective when the state space is small and discrete, but they scale poorly with state dimension. The number of state-action pairs grows exponentially with the dimension of the state representation, and tabular methods do not provide any mechanism for generalizing across unvisited states. This is impractical in the present setting, where the state is constructed from
a high-dimensional representation of the evolving numerical solution. DQN avoids this limitation by using a neural network, with weights shared across all inputs, to approximate the action value function. Because nearby states share the same network parameters, the learned function varies smoothly across the state space, which allows the network to produce reasonable value estimates for states not directly seen during training.\\
\end{remark}

As a value-based RL method, DQN uses a neural network to approximate an action value function that quantifies the expected long-term return associated with taking a particular action in a given state.  Toward this effect, rather than representing the policy $\pi$ directly, the RL agent learns the action value function
\begin{equation}
    Q^{\pi}(s,a) = \mathbb{E}_{\pi}\left[\sum_{t=0}^{\infty}\gamma^{t} r_{t} \,\middle|\, s_0 = s,\; a_0 = a\right],
\end{equation}
the expected return from taking action $a$ in state $s$ and following $\pi$ thereafter. It can be shown \cite{bellman1966dynamic} that the optimal action value function satisfies the Bellman optimality equation
\begin{equation} \label{eq:bellman}
    Q^{*}(s,a) = \mathbb{E}\left[r + \gamma \max_{a'} Q^{*}(s',a') \,\middle|\, s,a \right],
\end{equation}
from which a greedy policy $\pi^{*}(s) = \arg\max_{a} Q^{*}(s,a)$ is recovered. Since $\mathcal{A}$ is finite, this maximization is performed exhaustively: a single forward pass of the trained network on $s$ returns $Q(s,a)$ for every $a \in \mathcal{A}$, and the action with the largest value is selected.
Training follows the standard DQN procedure, in which the DQN is trained using the information collected as the agent interacts with the environment. To ensure that the agent explores a broad range of actions, actions are selected during training using $\epsilon$-greedy exploration: with probability $\epsilon$ a random action is selected from $\mathcal{A}$, and otherwise the greedy action with the largest predicted value is selected. The exploration parameter $\epsilon$ is decayed from $1.0$ to $0.05$ over the course of training. Each interaction produces a transition $(s,a,r,s')$, recording the current state $s$, the action $a$ taken by the agent, the resulting reward $r$, and the subsequent state $s'$. These transitions are stored in a replay buffer, which serves as a collection of the agent's past experiences. Rather than training the network using transitions in the order in which they occur, transitions are sampled randomly from this buffer in small groups, or minibatches. This reduces correlations between successive training samples and improves the stability of the learning process.

DQN employs two neural networks during training: an online network \(Q_\theta\), whose parameters are continually updated, and a target network \(Q_{\bar\theta}\), whose parameters are held fixed for a prescribed number of training steps. For each sampled transition, the target network is used to construct the regression target
\begin{equation} \label{eq:twonet}
    y = r + \gamma\,(1 - d)\, Q_{\bar{\theta}}\!\left(s', \arg\max_{a'} Q_{\theta}(s',a')\right),
\end{equation}
where the argmax is evaluated over the finite, discrete action set $\mathcal{A}$: a single
forward pass of the online network on $s'$ produces one value $Q_{\theta}(s', a')$ for each
candidate action $a' \in \mathcal{A}$, and the action having the largest such value is
selected. The first term in \eqref{eq:twonet} is the reward received for the action just taken, while the second estimates the future reward that can be obtained from the resulting state $s'$. If $s'$ is the terminal state, $d=1$ and the future reward contribution vanishes.  In \eqref{eq:twonet}, the online network $Q_\theta$ is used to identify the action expected to be best in the new state $s'$, while the target network $Q_{\bar\theta}$ evaluates the value of that action. Separating these two operations reduces the systematic overestimation of action values that can arise when the same network is used for both. The parameters $\theta$ of the online network are updated by minimizing the
Huber loss~\cite{huber1992robust} between its current prediction $Q_\theta(s,a)$
and the target value $y$. For a residual $\delta := Q_\theta(s,a) - y$, the
Huber loss is defined as
\begin{equation}
\mathcal{L}_\kappa(\delta) =
\begin{cases}
\tfrac{1}{2}\delta^2, & |\delta| \le \kappa, \\[4pt]
\kappa\left(|\delta| - \tfrac{1}{2}\kappa\right), & |\delta| > \kappa,
\end{cases}
\end{equation}
where $\kappa > 0$ is a threshold parameter.
The Huber loss is selected because it is less sensitive than a squared loss to occasional large prediction errors during training.
Through repeated updates using transitions sampled from the replay buffer, the network learns to estimate the expected long-term return associated with each available action. During deployment, no exploration is performed: the policy is purely greedy, and the learned action value function is used to select the action with the largest predicted return for the current state. The specific application of this DQN framework to adaptive model selection and DD within O-SAM is described next, in Section~\ref{sec:rl-osam}.


\subsection{RL-based online adaptation of O-SAM} \label{sec:rl-osam}

We now specialize the RL framework described above to the adaptive O-SAM problem. Recall from Section \ref{sec:o-sam} that, for a given DD and model assignment, the O-SAM algorithm advances the solution by solving the subdomain-local problems and iteratively exchanging Dirichlet data until convergence is declared. Once the Schwarz iteration has converged, the coupled solution is advanced forward in time. In the present approach, RL does not modify the Schwarz iteration itself, but instead selects the subdomain models -- and, possibly, the domain decomposition -- used by O-SAM in each decision window.

\begin{figure}[ht!]
\centering
\includegraphics[width=\textwidth]{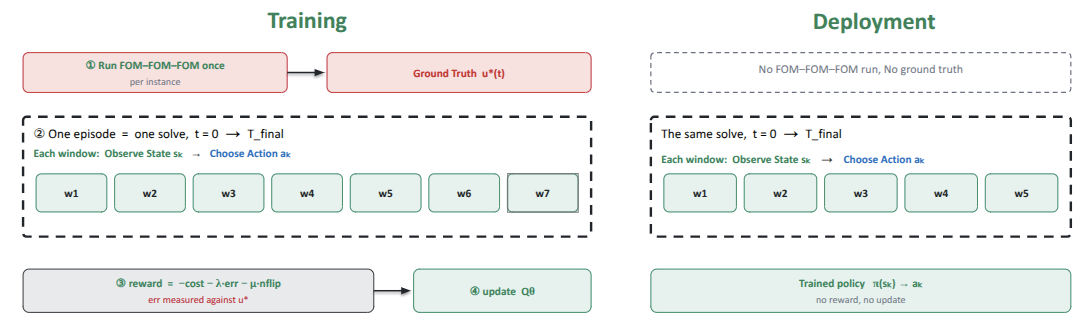}
\caption{Offline training and online deployment of the RL–O-SAM framework. During training (left), a high-fidelity reference solution $u^*(t)$ is computed for each problem instance and used to evaluate the error term in the reward \eqref{eq:reward}. Each episode advances from $t=0$ to $T_{\mathrm{final}}$, with the reward evaluated after each decision window and used to train the DQN. During deployment (right), neither the reference solution nor the reward is required; the trained policy selects an action from the observed state alone.}
\label{fig:train-vs-deploy}
\end{figure}

\begin{figure}[htb]
\centering
\includegraphics[width=\textwidth]{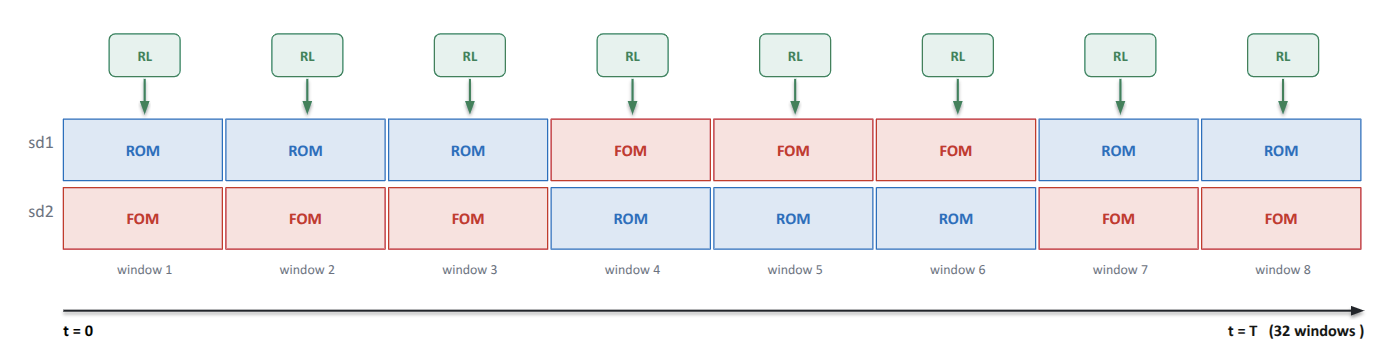}
\caption{RL–O-SAM workflow for two subdomains, showing the first eight of the 32 decision windows of an episode. The episode advances from left to right in time. At the beginning of each window, the agent observes the current state and selects a FOM/ROM assignment, which is held fixed while O-SAM advances the coupled solution over the window. The agent therefore acts once per decision window, rather than at each Schwarz iteration or time step (see Figure \ref{fig:train-vs-deploy}); the O-SAM coupling remains unchanged, while the models assigned to the subdomains may vary.}
\label{fig:rl-schwarz}
\end{figure}

Suppose that the time interval $[0,T_{\text{final}}]$ associated with one \textit{simulated trajectory}, i.e., one complete simulation of a particular problem instance, defined by a given choice of initial conditions, boundary conditions and/or physical parameters, is divided into a prescribed number of \textit{decision windows}, as illustrated in Figure \ref{fig:rl-schwarz}.  We refer to this simulated trajectory as an \textit{episode}. The proposed RL--O-SAM workflow consists of two stages: an \textit{offline training stage} and an \textit{online deployment stage}, as illustrated in Figure \ref{fig:train-vs-deploy}.

\subsubsection{Offline training stage} \label{sec:offline}

In the offline training stage, we first define the discrete action space $\mathcal{A}$ available to the RL agent and construct the corresponding subdomain-local ROMs. We consider two approaches for defining $\mathcal{A}$.

In the first approach, termed ``fixed-DD", a fixed overlapping decomposition of the computational domain into $n_d$ subdomains is prescribed. To generate training data for the subdomain-local OpInf ROMs, we perform all-FOM simulations in which the subdomain-local FOMs are coupled using O-SAM, for a collection of problem instances characterized by different initial conditions, boundary conditions, and/or physical parameters. The resulting subdomain-local solution snapshots and Schwarz boundary data are then used to construct a pre-trained OpInf ROM for each subdomain, following the procedure described in Section \ref{sec:opinf-osam}. Thus, both a FOM (denoted by `F') and a pre-trained OpInf ROM (denoted by `R') are available in each subdomain. The action space $\mathcal{A}$ now consists of all possible FOM/ROM assignments across the $n_d$ subdomains, giving a total of $|\mathcal{A}|=2^{n_d}$ possible actions. For example, for $n_d=3$, $\mathcal{A}=\{\text{FFF, FFR, FRF, FRR, RFF, RFR, RRF, RRR}\}$, where the position of each `F' or `R' identifies the model assigned to the corresponding subdomain.

In the second approach, termed ``adaptive-DD", both the DD and the FOM/ROM subdomain assignment are included in the action space $\mathcal{A}$. Rather than prescribing a single fixed decomposition, we define a discrete set of $N_{\mathrm{DD}}$ candidate overlapping decompositions of the computational domain, each consisting of $n_d$ subdomains. For example, for the one-dimensional (1D) domain $\Omega=[0,1]$ and $n_d=2$, three candidate decompositions could be prescribed as
\begin{equation} \begin{aligned} \mathcal{P}_1 &: \quad [0,0.5]\cup[0.4,1],\\ \mathcal{P}_2 &: \quad [0,0.3]\cup[0.2,1],\\ \mathcal{P}_3 &: \quad [0,0.8]\cup[0.5,1]. \end{aligned} 
\end{equation}
For each candidate decomposition, every subdomain may be assigned either a FOM (`F') or a pre-trained OpInf ROM (`R'). Thus, an action specifies both the candidate decomposition, and the FOM/ROM assignment on that decomposition. The resulting action space will contain $ |\mathcal{A}|=N_{\mathrm{DD}}2^{n_d} $ possible actions. For the example above, $N_{\mathrm{DD}}=3$ and $n_d=2$, yielding $3\times2^2=12$ possible actions. The action $(\mathcal{P}_2, \text{FR})$, for instance, selects decomposition $\mathcal{P}_2$, with a FOM employed on its first subdomain and a ROM on its second subdomain.

Once the action space and all required subdomain-local ROMs have been constructed, the RL agent is trained offline using a collection of problem instances characterized by different initial conditions, boundary conditions, and/or physical parameters. Each problem instance defines an episode, comprised of prescribed decision windows, as illustrated in the left panel of Figure \ref{fig:train-vs-deploy}. At the beginning of each decision window, the agent observes the current state \(s_k\) and selects an action \(a_k\in\mathcal{A}\). The selected action determines the FOM/ROM assignment across the subdomains and, when adaptive-DD is considered, the DD to be used. This configuration is held fixed throughout the decision window while the coupled problem is advanced using O-SAM. At the end of each decision window, the agent receives a reward that balances computational efficiency, solution accuracy, and the cost of changing the selected configuration. Specifically, we define the reward as
\begin{equation} \label{eq:reward}
 r_k=-\mathrm{cost}_k-\alpha\,\mathrm{err}_k-\beta\,n_{\mathrm{flip},k}, 
 \end{equation}
where $\mathrm{cost}_k$ denotes the computational cost incurred during decision window $k$, $\mathrm{err}_k$ denotes the error in the coupled solution relative to a reference solution, and $n_{\mathrm{flip},k}$ penalizes changes in the selected action between consecutive decision windows. Note that the reward is evaluated from the hybrid solution actually computed with the selected configuration during training; hence, the only reference solution required is the all-FOM solution, and no simulations of the individual FOM/ROM combinations need to be computed in advance. Similarly, the subdomain-local OpInf ROMs are constructed from all-FOM snapshots only. 
The precise definition of $\mathrm{err}_k$ is problem-dependent and is given for the advection-diffusion and clamped linear elastic wave propagation benchmarks in Sections \ref{sec:advdiff} and \ref{sec:clamped}, respectively.
The parameters $\alpha$ and $\beta$ in \eqref{eq:reward} control the relative importance of solution accuracy and switching, respectively. Since $\mathrm{cost}_k$ is measured as a wall time, while $\mathrm{err}_k$ is a relative error and $n_{\mathrm{flip},k}$ is a count, $\alpha$ and $\beta$ carry the non-dimensionalization of \eqref{eq:reward} and have units of time. They therefore have a direct interpretation: $\alpha$ is the computational time one is willing to spend to reduce the relative error by one unit, and $\beta$ is the time-equivalent penalty assigned to each model switch. During offline RL training, a high-fidelity reference solution $u^*(t)$ is computed for each training problem instance and used to evaluate $\mathrm{err}_k$. The specific reference solution employed depends on the test case and is defined in Section \ref{sec:results}. The resulting transition $(s_k,a_k,r_k,s_{k+1})$ is used to train the DQN according to the procedure described in Section \ref{sec:dqn}. 

\subsubsection{Online deployment stage} \label{sec:online}

Having described the offline training stage of our RL algorithm, we now move on to the online deployment stage, depicted in Figure \ref{fig:rl-schwarz}. Once offline training is complete, the trained DQN policy and all pre-trained subdomain-local OpInf ROMs are held fixed and deployed for the online simulation of a new problem instance. At the beginning of each decision window, the RL agent observes the current state $s_k\in\mathcal{S}$, which consists of a fixed-dimensional representation of the current coupled solution together with the action $a_{k-1}$ selected in the preceding decision window. Including the previous action provides the agent with information about the current FOM/ROM assignment and, in the adaptive-DD setting, the current domain decomposition. The specific representation of the state employed in the numerical examples is described in Section \ref{sec:results}.

Given the state \(s_k\), the trained DQN selects an action \(a_k\in\mathcal{A}\). This action determines the FOM/ROM assignment on each subdomain and, in the adaptive-DD setting, the DD to be employed during the current decision window. The selected configuration is held fixed throughout the decision window while the coupled problem is advanced using O-SAM. In particular, the RL agent does not act at individual time steps or Schwarz iterations. At the beginning of the subsequent decision window, a new state is constructed from the updated coupled solution and the preceding action, and the process is repeated until the final simulation time \(T\) is reached.

In contrast to the offline training stage, online deployment requires neither a reference solution nor evaluation of the reward function. The trained DQN selects actions based solely on the observed state, and neither the DQN nor the pre-trained OpInf ROMs are updated during the online simulation. Thus, the high-fidelity reference solution used to evaluate the error contribution to the reward during training is not required for a new problem instance at deployment.  Importantly, the trained policy is deployed predictively on problem instances not encountered during training. As demonstrated in Section \ref{sec:results}, the numerical examples considered herein train the policy using one set of initial conditions and evaluate the resulting policy on initial conditions withheld from RL training, thereby assessing the ability of the learned policy to generalize beyond the problem instances used to train it. Whether the underlying OpInf ROMs are themselves deployed predictively or reproductively depends on the benchmark and is specified in Section \ref{sec:results}.\\
\begin{remark}
 It is interesting to observe that the adaptive O-SAM framework may also be interpreted from a space-time DD perspective. Within each decision window, O-SAM performs an overlapping DD in space, with the subdomain-local models coupled through Schwarz transmission conditions. The successive decision windows, in contrast, define a non-overlapping decomposition of the time interval. Thus, the overall adaptive simulation can be viewed as a space-time decomposition consisting of overlapping Schwarz subdomains in space and non-overlapping subdomains in time. This interpretation is particularly apparent in the space-time representation of the advection-diffusion example in Figure \ref{fig:advdiff-fixed-dec}.
\end{remark}

\section{Numerical results} \label{sec:results}

We evaluate the proposed RL--O-SAM framework on two time-dependent benchmark problems: a 1D transient advection-diffusion problem with a moving front (Section \ref{sec:advdiff}), and a linear elastic wave propagation problem in solid mechanics (Section \ref{sec:clamped}). These examples are used to assess the ability of the learned policy to perform online FOM-ROM switching as the solution evolves. For the advection-diffusion problem, we consider both a fixed-DD case, in which the RL agent selects only the FOM/ROM assignment, as well as an adaptive-DD case, in which the agent selects both the DD and the corresponding FOM/ROM assignment.

For all numerical experiments reported below, the DQN $Q_\theta$ (see Section \ref{sec:dqn}) is represented by a multilayer perceptron (MLP) with two hidden layers of 256 units and ReLU activation functions, with one output for each available action. Training is performed using the Adam optimizer with a learning rate of $10^{-3}$, a replay buffer containing up to $2\times10^4$ transitions, minibatches of 128 transitions, and a discount factor of $\gamma=0.99$. The target network is synchronized with the online network every 400 environment steps. Problem-specific definitions of the state, action space, reward terms, and associated parameters are provided within the corresponding numerical examples.

The same Schwarz solver settings are used throughout the numerical experiments. Schwarz convergence is declared when the change in the interface traces between successive iterations, measured in the discrete $L^2$ norm over the decision window and accumulated over all interfaces, falls below an absolute tolerance of $10^{-6}$. A maximum of 25 Schwarz iterations is permitted per decision window.

\subsection{Advection-diffusion benchmark} \label{sec:advdiff}

We first assess the learned policy on a 1D transient advection-diffusion problem,
\begin{equation}
    u_t + c\,u_x = \nu\,u_{xx}, \qquad x \in (0,1), \quad t \in [0,T_{\text{final}}],
\end{equation}
with $c = 1$, $\nu = 10^{-2}$ and $T_{\text{final}} = 0.6$, which gives rise to a  P\'eclet number of $cL/\nu = 100$. A Dirichlet condition $u(0,t)=1$ is imposed at the inflow boundary, together with a homogeneous Neumann condition at the outflow. The initial condition is given by the smoothed front
\begin{equation} \label{eq:ic}
    u(x,0) = \frac{1}{2}\left[1 - \tanh\!\left(\frac{x - x_0}{\delta}\right)\right], \qquad \delta = 0.01,
\end{equation}
where $x_0$ determines the initial location of the front. During RL training, $x_0$ is sampled uniformly from $[0.05,0.5]$, so that each episode corresponds to a different problem instance. As the sharp front propagates across the domain, different subdomains require higher-fidelity resolution at different times, motivating adaptive FOM/ROM model selection.

The domain $\Omega=(0,1)$ is first discretized using \(N=2048\) interior nodes. Because conformal subdomain meshes are employed, the overlapping subdomain meshes are obtained by restricting this global discretization to the corresponding subdomain intervals. The precise subdomain locations depend on whether a fixed- or adaptive-DD is employed and are specified below. Second-order central differences are used for diffusion and first-order upwinding for advection.  The resulting semi-discrete systems are integrated using an adaptive Runge-Kutta method. The subdomains are coupled using multiplicative O-SAM with Dirichlet transmission conditions, as discussed in Section \ref{sec:o-sam}.

The subdomain-local OpInf ROMs utilized within our couplings are constructed offline using 10 FOM trajectories, each corresponding to a different initial condition \eqref{eq:ic}  with
\begin{equation} \label{eq:x0}
    x_0\in\{0.280,\,0.478,\,0.115,\,0.477,\,0.190,\,0.240,\,0.423,\,0.234,\,0.297,\,0.062\}.
    \end{equation}
Each trajectory contains 401 snapshots sampled uniformly over $[0,T_{\text{final}}]$. The same 10 trajectories are used to construct all subdomain-local ROMs, with the snapshots restricted to the degrees of freedom associated with each subdomain. Thus, the local bases are constructed from a common snapshot set rather than from separate FOM simulations.

\begin{figure}[htbp]
\centering
\includegraphics[width=0.62\textwidth]{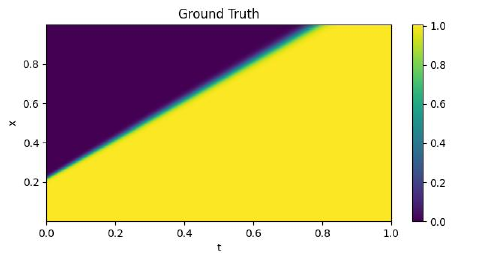}
\caption{Monolithic full order solution in the space-time plane, used as the ground truth for both settings below. The front enters at the inflow boundary and advects to the right at speed $c$, remaining sharp over the whole time interval.}
\label{fig:advdiff-truth}
\end{figure}

For the advection-diffusion benchmark, the state supplied to the DQN consists of a fixed-dimensional representation of the current solution $u$ together with the action selected in the preceding decision window. The $N=2048$-dimensional solution field is max-pooled to 200 components to reduce the dimension of the solution state provided to the DQN, and the previous action is represented using one-hot encoding, i.e., a binary vector with a single entry equal to one at the index corresponding to the selected action and zeros elsewhere.  For the reward function \eqref{eq:reward}, $\mathrm{cost}_k$ is taken to be the wall time required to advance decision window $k$, while $\mathrm{err}_k$ is the relative error with respect to the corresponding monolithic FOM solution.  To evaluate this error, the converged subdomain solutions are first assembled into a global solution field \(\mathbf{u}_k\) on the global grid, with the solutions from neighboring subdomains averaged at grid points in the overlap regions. The error is then computed as
\begin{equation}
\mathrm{err}_k = \frac{\|~u_k-~u_k^*\|_2} {\|~u_k^*\|_2}, 
\end{equation}
where $\mathbf{u}_k^*$ denotes the corresponding monolithic FOM solution on the global grid and $\|\cdot\|_2$ denotes the discrete $\ell_2$ norm.
We set $\alpha=500$ and $\beta=1.5$. These values were fixed prior to training by sweeping each parameter over several orders of magnitude and selecting values for which no single term (cost, accuracy, or switching) dominates the reward; the model assignments are then determined by the learned policy. The switching term $n_{\mathrm{flip},k}$ counts the number of subdomains whose FOM/ROM assignment changes relative to the preceding decision window; in the adaptive-DD setting (Section \ref{sec:adaptive_sd}), a change in the selected decomposition is also penalized. 
The purpose of this term is to discourage unnecessary changes in the model assignment or DD between consecutive decision windows, which may incur additional computational overhead and lead to frequent oscillations between otherwise comparable actions.
No switching penalty is applied during the first decision window.  The DQN is trained for 500 episodes and subsequently evaluated predictively for $x_0=0.22$. 
Since this initial condition is not used during either DQN training or OpInf ROM construction, both the learned policy and the underlying OpInf ROMs are evaluated predictively for this test case.
The monolithic FOM solution for $x_0=0.22$, shown in Figure \ref{fig:advdiff-truth}, is used solely as a reference for evaluating the accuracy of the deployed policy.

We consider two configurations. In the first fixed-DD case,  the DD is fixed and the agent selects only the FOM/ROM assignment among the three subdomains, yielding $2^3=8$ possible actions (Section \ref{sec:fixed_sd}). In the second adaptive-DD case, the agent selects both the domain decomposition and the FOM/ROM assignment, yielding 40 possible actions (Section \ref{sec:adaptive_sd}).

\subsubsection{Fixed-DD case} \label{sec:fixed_sd}

First, we consider the first offline training approach described in Section \ref{sec:rl-osam}, in which the DD is fixed, and we are learning a policy that determines online whether each subdomain is assigned a FOM (`F') or a ROM ('R').  
In this study, the domain is decomposed into three overlapping subdomains of equal length with 30\% overlap,
\begin{equation}
 \Omega_1=[0,0.4167],\qquad \Omega_2=[0.2917,0.7083],\qquad \Omega_3=[0.5833,1], 
 \end{equation}
containing 853, 854, and 853 of the 2048 nodes, respectively. Each subdomain is equipped with an OpInf ROM of dimension $r=12$. The ROM dimension was selected by testing a range of candidate values, with $r=12$ being the smallest dimension that provided acceptable ROM accuracy for this configuration. The OpInf least-squares problem  \eqref{eq:opinf-learning} uses a Tikhonov regularization with $\eta = 10^{-8}$.  
For the RL training, the time interval $[0,T_{\text{final}}]$ is divided uniformly into 128 decision windows. At the beginning of each window, the agent selects one of the $2^3=8$ possible FOM/ROM assignments, which is held fixed while O-SAM advances the coupled solution over the window. After the Schwarz iteration converges, the RL agent observes the resulting state and selects the assignment for the next decision window.

\begin{table}[ht!]
\centering
\begin{tabular}{lcccc}
\hline
policy & relative $L^2$ error & Schwarz iterations & wall time [s] & reward \\
\hline
    RL (8 actions) & \textbf{$8.02\times 10^{-4}$} & \textbf{$3.00$} & \textbf{$1.91$} & \textbf{$\mathbf{-67.4}$} \\
FFF            &  $\mathbf{1.19\times 10^{-4}}$ & $\mathbf{1.61}$ & $3.23$ & $-106.9$ \\
FFR            & $4.19\times 10^{-3}$ & $2.11$ & $1.62$ & $-141.7$ \\ 
RFF            & $8.31\times 10^{-3}$ & $2.34$ & $2.66$ & $-310.1$ \\ 
RFR            & $9.19\times 10^{-3}$ & $2.45$ & $1.93$ & $-325.5$ \\ 
FRR            & $9.50\times 10^{-3}$ & $3.04$       & $1.65$ & $-333.4$ \\ 
FRF            & $9.65\times 10^{-3}$ & $3.97$       & $3.47$ & $-356.1$ \\ 
RRF            & $1.17\times 10^{-2}$ & $2.89$       & $2.76$ & $-419.9$ \\ 
RRR            & $1.18\times 10^{-2}$ & $4.67$ & $\mathbf{0.51}$ & $-397.9$ \\
\hline
\end{tabular}
\caption{Advection-diffusion benchmark, fixed-DD. Results for the learned policy and the eight possible fixed FOM/ROM assignments. `F' and `R' denote full order and reduced order subdomain models, respectively, ordered from left to right. The relative $L^2$ error is computed with respect to the monolithic FOM reference solution. `Schwarz iterations' denotes the mean number of multiplicative Schwarz iterations per decision window, averaged over the 128 decision windows. `Reward' denotes the sum of the rewards over all decision windows of the episode.  The best value in each column is shown in bold.}
\label{tab:advdiff8}
\end{table}

Table \ref{tab:advdiff8} compares the trained policy with each of the eight fixed FOM/ROM assignments. The learned policy achieves a relative $L^2$ error of $8.02\times10^{-4}$, approximately five times smaller than that of the most accurate fixed assignment containing at least one ROM (FFR), while requiring an average of 3.00 Schwarz iterations per decision window, compared with 1.61 for the all-FOM assignment. The increase in Schwarz iterations when ROMs are introduced is consistent with the observations in Wentland et al. \cite{Wentland2025Schwarz}, where Schwarz couplings involving lower-dimensional  ROMs were found to generally require more iterations to converge than the corresponding all-FOM coupling. This behavior arises because inaccuracies in the ROM solution can lead to larger discrepancies between neighboring subdomain solutions, requiring the Schwarz iteration to work harder to reconcile the solutions across the overlap.  Despite the observed increase in Schwarz iterations, the learned policy reduces the wall time from $3.23$ s for FFF to $1.91$ s.  As expected, the all-ROM (RRR) coupling yields the lowest wall time, $0.51$ s, but at the expense of a substantially larger relative $L^2$ error of $1.18\times10^{-2}$. Thus, the learned policy provides a compromise between the high accuracy of the all-FOM assignment and the low computational cost of the all-ROM assignment.

\begin{figure}[ht!]
\centering
\includegraphics[width=0.62\textwidth]{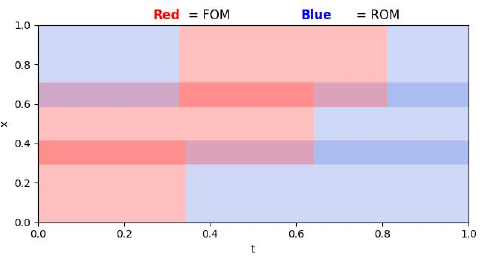}
\caption{Advection-diffusion benchmark, fixed-DD. FOM/ROM model assignment selected by the learned policy for each subdomain as a function of time, with red denoting the FOM and blue denoting the ROM. The FOM assignment shifts from left to right as the sharp front advects through the domain, preferentially providing full order resolution in the subdomain containing the moving front.}
\label{fig:advdiff-fixed-dec}
\end{figure}

\begin{figure}[ht!]
\centering
\includegraphics[width=\textwidth]{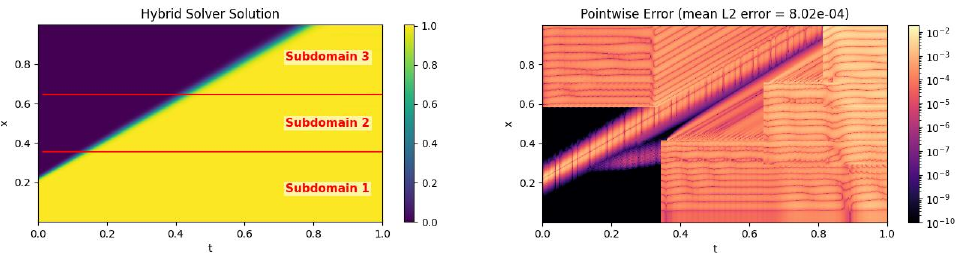}
\caption{Advection-diffusion benchmark, fixed-DD. \textit{Left:} hybrid solution produced by the learned policy, with the two subdomain interfaces indicated in red. \textit{Right:} pointwise error relative to the monolithic FOM reference solution, shown on a logarithmic scale. The error is largest in the reduced order subdomains and near the advecting front, and smallest where the policy selects the FOM.}
\label{fig:advdiff-fixed-solerr}
\end{figure}

As shown in Figure \ref{fig:advdiff-fixed-dec}, the learned policy adapts the model assignment as the sharp front propagates through the domain, preferentially assigning the FOM to the subdomain containing the front while using ROMs elsewhere. As the front advects from left to right, the FOM assignment correspondingly shifts from one subdomain to the next. Figure \ref{fig:advdiff-fixed-solerr} shows the resulting hybrid solution and its pointwise error relative to the monolithic FOM reference solution.

\begin{figure}[ht!]
\centering
\includegraphics[width=0.52\textwidth]{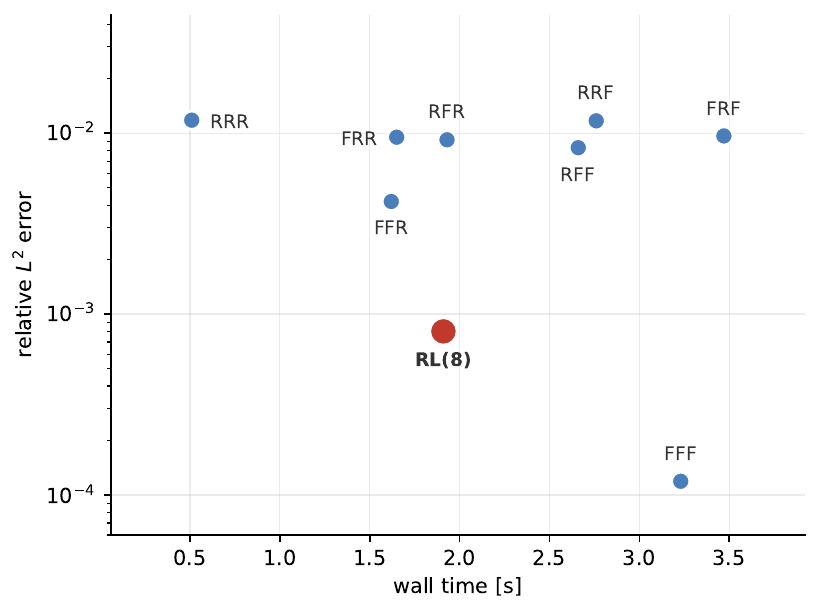}
\caption{Advection-diffusion benchmark, fixed-DD.  Pareto plot showing the relative $L^2$ error vs. measured wall time for the learned policy and all fixed FOM/ROM assignments. The learned policy, denoted by `RL(8)', is shown in red. Policies farther toward the lower left provide a more favorable accuracy-cost tradeoff, making the learned policy Pareto-optimal among the fixed assignments considered.}
\label{fig:advdiff-reward}
\end{figure}

Finally, the accuracy-cost tradeoff is summarized by the Pareto plot in Figure \ref{fig:advdiff-reward}, which shows the relative $L^2$ error vs. wall time for the learned policy and all fixed model assignments. It is clear from this figure that the learned policy, denoted by `RL(8)', is Pareto-optimal among the fixed assignments considered.\\

 \textbf{4.1.1.1.  Effect of the switching penalty.  } 
\begin{figure}[ht!]
\centering
\includegraphics[width=0.72\textwidth]{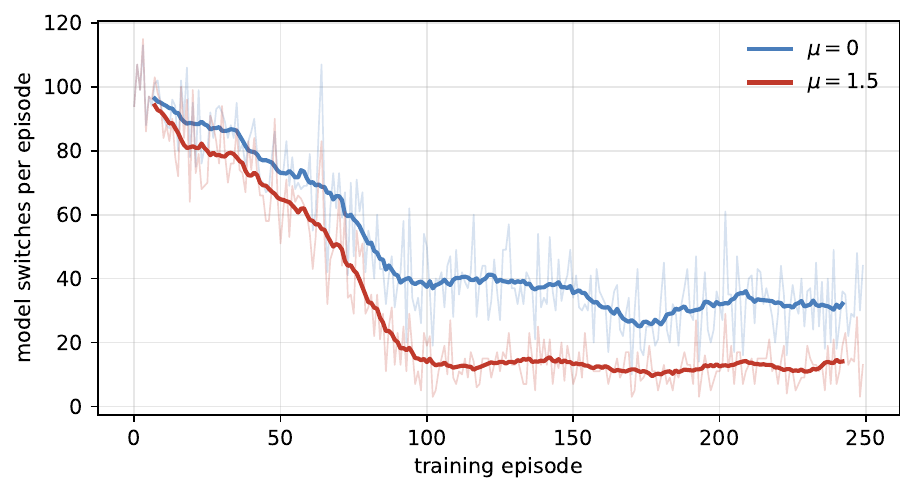}
\caption{Advection–diffusion benchmark, fixed-DD, effect of the switching penalty during training. Number of FOM/ROM model switches per training episode for $\beta=0$ (no switching penalty) and $\beta=1.5$ (including a switching penalty). Faint lines show the number of switches in each episode, while solid lines show the corresponding moving averages. Without the switching penalty, the number of switches levels off at approximately 33 per episode, whereas with $\beta=1.5$, it decreases to approximately 13 per episode.}
\label{fig:advdiff-mu-flips}
\end{figure}
It is interesting to understand the effect of the switching penalty in the reward function \eqref{eq:reward}, in particular, the extent to which it discourages unnecessary changes in the FOM/ROM assignment. To isolate the effect of this term, we retrain the fixed-DD policy with $\beta=0$ and $\beta=1.5$ in \eqref{eq:reward}, while holding all other parameters fixed. Figure \ref{fig:advdiff-mu-flips} shows the number of model switches per episode during training. Both policies initially exhibit approximately 100 switches per episode, as the untrained agent explores the action space. As training progresses, the number of switches decreases for both policies. Without the switching penalty ($\beta=0$), however, the number eventually levels off at approximately 33 switches per episode, whereas with $\beta=1.5$, it continues to decrease to approximately 13.

\begin{figure}[ht!]
\centering
\includegraphics[width=\textwidth]{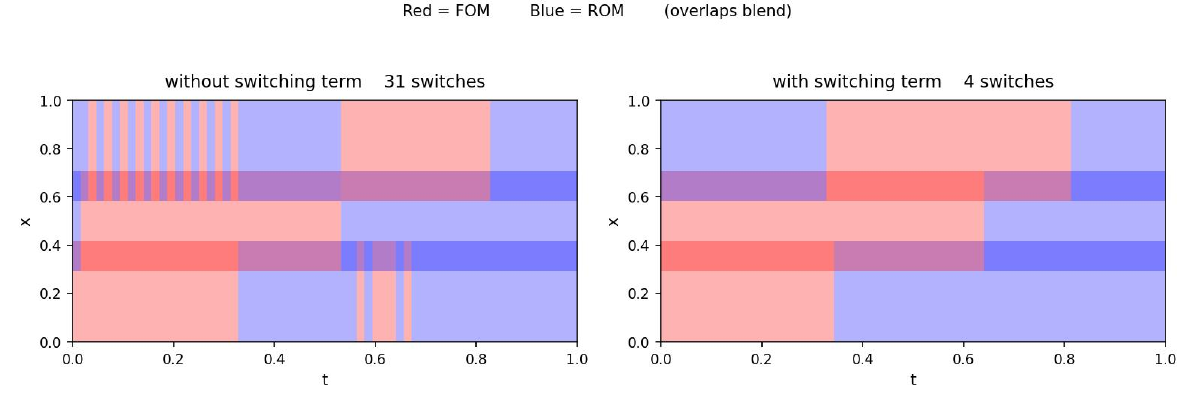}
\caption{Advection-diffusion benchmark, fixed-DD, effect of the switching penalty on the deployed policy. FOM/ROM model assignments selected by policies trained without the switching penalty ($\beta=0$, left) and with the switching penalty ($\beta=1.5$, right), with red denoting the FOM and blue denoting the ROM. The switching penalty suppresses the rapid FOM/ROM switching observed for $\beta=0$, reducing the total number of model switches over the episode from 31 to 4.}
\label{fig:advdiff-mu-decision}
\end{figure}

 The effect of the switching penalty is even more apparent when examining the deployed policies. As shown in Figure \ref{fig:advdiff-mu-decision}, the policy trained with $\beta=0$ (no switching penalty) switches models 31 times over the episode, with much of this switching occurring as rapid back-and-forth changes early in the simulation. In contrast, the policy trained with $\beta=1.5$ switches only four times, producing a considerably more coherent model-assignment pattern in which the FOM tracks the advecting front before transitioning to an all-ROM assignment after the front leaves the domain.

The above results reinforce the fact that the purpose of the switching penalty is not to improve solution accuracy, but rather to regularize the behavior of the learned policy by suppressing unnecessary FOM/ROM switching. This is particularly important for deployment within a production code, where each model switch may require a state transfer between the FOM and ROM representations and therefore incur additional computational costs.

\subsubsection{Adaptive-DD case} \label{sec:adaptive_sd}

Next, we assess the second training strategy introduced in Section \ref{sec:adapt-rl}, in which the RL agent selects both the DD and the FOM/ROM assignment. Five candidate DDs, denoted by \(\mathcal{P}_i\), \(i=1,\ldots,5\), are considered. These decompositions are obtained by varying the locations of the two interior interfaces while maintaining a 25\% overlap between adjacent subdomains. The resulting subdomain extents and node counts are listed in Table \ref{tab:advdiff-partitions}.
\begin{table}[ht!]
\centering
\begin{tabular}{c|ccc|ccc}
\hline
          & \multicolumn{3}{c}{subdomain extent} & \multicolumn{3}{c}{\# nodes} \\
partition & left & middle & right & left & middle & right \\
\hline
$\mathcal{P}_1$ & $[0,\,0.551]$ & $[0.484,\,0.811]$ & $[0.757,\,1]$ & $1128$ & $670$ & $497$  \\
$\mathcal{P}_2$ & $[0,\,0.454]$ & $[0.377,\,0.758]$ & $[0.688,\,1]$ & $929$  & $780$ & $638$  \\
$\mathcal{P}_3$ & $[0,\,0.375]$ & $[0.292,\,0.708]$ & $[0.625,\,1]$ & $768$  & $854$ & $768$  \\
$\mathcal{P}_4$ & $[0,\,0.285]$ & $[0.222,\,0.646]$ & $[0.559,\,1]$ & $583$  & $869$ & $904$  \\
$\mathcal{P}_5$ & $[0,\,0.194]$ & $[0.151,\,0.566]$ & $[0.479,\,1]$ & $397$  & $851$ & $1067$ \\
\hline
\end{tabular}
\caption{Advection-diffusion benchmark, adaptive-DD. Subdomain extents and node counts for the five candidate DDs $\mathcal{P}_i$, $i=1,\ldots,5$ available to the RL agent. Each decomposition consists of three overlapping subdomains with 25\% overlap between adjacent subdomains. A separate OpInf ROM is constructed offline for each of the 15 subdomains, so that all candidate ROMs are available prior to online deployment.}
\label{tab:advdiff-partitions}
\end{table}
The set of candidate domain decompositions is prescribed offline, and a separate OpInf ROM is constructed offline for each subdomain of each $\mathcal{P}_i$, resulting in a total of 15 subdomain-local ROMs. Thus, all ROMs that may be selected by the agent are available prior to deployment, and no ROM construction is performed online. For this configuration, each ROM has dimension $r=10$, selected as the smallest dimension among the candidate values tested that provided acceptable ROM accuracy. The OpInf least-squares problem \eqref{eq:opinf-learning} utilizes a Tikhonov regularization parameter of $\eta=10^{-8}$.  
At the beginning of each decision window, the agent selects one of the five candidate domain decompositions $\mathcal{P}_i$ together with one of the eight possible FOM/ROM assignments, yielding an action space of $5\times 2^3=40$ actions. The selected decomposition and FOM/ROM assignment are held fixed throughout the decision window while O-SAM advances the coupled solution.

\begin{table}[ht!]
\centering
\begin{tabular}{lcccc}
\hline
policy & relative $L^2$ error & Schwarz iterations & wall time [s] & reward \\
\hline
RL (40 actions) & $1.03\times 10^{-3}$ & $2.73$ & $2.84$ & $\mathbf{-171.2}$ \\
FFF             & $\mathbf{1.23\times 10^{-4}}$ & $\mathbf{1.60}$ & $3.64$ & $-205.5$ \\
FFR             & $7.23\times 10^{-3}$ & $1.81$       & $3.25$ & $-405.2$ \\ 
RFF             & $1.30\times 10^{-2}$ & $1.95$       & $3.70$ & $-945.5$ \\ 
RFR             & $1.45\times 10^{-2}$ & $2.20$       & $2.93$ & $-1027.6$ \\ 
FRR             & $1.54\times 10^{-2}$ & $2.13$      & $2.36$ & $-1028.0$ \\ 
FRF             & $1.58\times 10^{-2}$ & $2.34$      & $4.79$ & $-1091.3$ \\ 
RRR             & $1.81\times 10^{-2}$ & $3.94$ & $\mathbf{0.39}$ & $-1226.3$ \\
RRF             & $1.82\times 10^{-2}$ & $3.19$       & $3.74$ & $-1286.5$ \\ 
\hline
\end{tabular}
\caption{Advection-diffusion benchmark, adaptive-DD. Results for the learned policy with an action space comprising 40 actions, obtained by combining the five candidate domain decompositions $\mathcal{P}_i$, $i=1,\ldots,5$ listed in Table \ref{tab:advdiff8} with the eight possible FOM/ROM assignments. `F' and `R' denote full order and reduced order subdomain models, respectively, ordered from left to right. The relative $L^2$ error is computed with respect to the monolithic FOM reference solution. `Schwarz iterations' denotes the mean number of multiplicative Schwarz iterations per decision window, averaged over the 128 decision windows. `Reward' denotes the sum of the rewards over all decision windows of the episode. The best value in each column is shown in bold.}
\label{tab:advdiff40}
\end{table}

Table \ref{tab:advdiff40} summarizes the performance of the learned policy with an action space comprising 40 possible actions, obtained by combining the five candidate DDs $\mathcal{P}_i$, $i=1,\ldots,5$, with the eight possible FOM/ROM assignments. The learned policy achieves a relative $L^2$ error of $1.03\times10^{-3}$, making it approximately seven times more accurate than the FFR coupling, 
the most accurate baseline FOM/ROM assignment containing at least one ROM.  The learned policy requires an average of 2.73 Schwarz iterations per decision window, compared with 1.60 for the all-FOM (FFF) assignment. 
Although an individual ROM subdomain solve is approximately 34-44 times less expensive than the corresponding FOM solve, it is interesting to observe that this reduction does not translate directly into an equivalent reduction in overall wall time. The increased number of Schwarz iterations associated with FOM-ROM coupling partially offsets the computational savings provided by the reduced order subdomain solves.
Despite requiring more Schwarz iterations, the learned policy has a lower wall time than FFF, $2.84$ s compared with $3.64$ s. The all-ROM (RRR) assignment has the lowest wall time, $0.39$ s, but its relative $L^2$ error, $1.81\times10^{-2}$, is approximately 18 times larger than that of the learned policy.

\begin{figure}[ht!]
\centering
\includegraphics[width=0.62\textwidth]{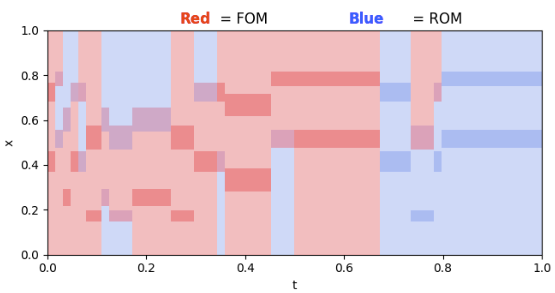}
\caption{Advection-diffusion benchmark, adaptive-DD. Domain decomposition and FOM/ROM model assignment selected by the learned policy as functions of time, with red denoting the FOM and blue denoting the ROM. Changes in the subdomain boundaries indicate changes in the selected candidate decomposition $\mathcal{P}_i$, while changes in color indicate changes in the FOM/ROM assignment.}
\label{fig:advdiff-adaptive-dec}
\end{figure}

\begin{figure}[ht!]
\centering
\includegraphics[width=0.49\textwidth]{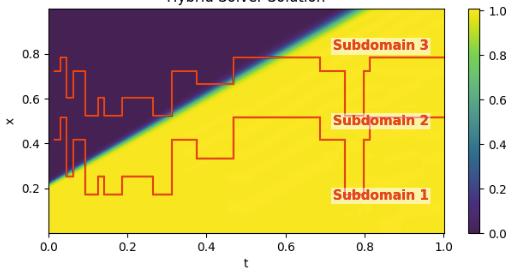}
\hfill
\includegraphics[width=0.49\textwidth]{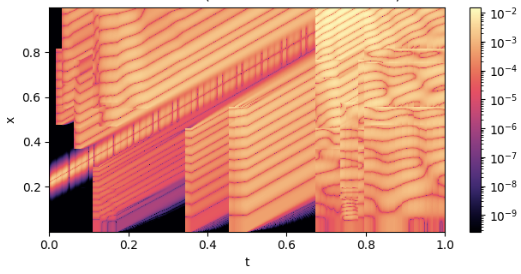}
\caption{Advection-diffusion benchmark, adaptive-DD. \textit{Left:} hybrid solution produced by the learned policy, with the selected subdomain interfaces shown in red and the three subdomains labeled. The interface locations change between decision windows as the agent selects among the five candidate domain decompositions $\mathcal{P}_i$. \textit{Right:} pointwise error relative to the monolithic FOM reference solution, shown on a logarithmic scale. Changes in the selected domain decomposition are visible as discontinuities in the interface locations.}
\label{fig:advdiff-adaptive-solerr}
\end{figure}

Figures \ref{fig:advdiff-adaptive-dec} and \ref{fig:advdiff-adaptive-solerr} illustrate how the learned policy, `RL(40)', jointly adapts the DD and FOM/ROM assignment as the solution evolves. Figure \ref{fig:advdiff-adaptive-dec} shows the corresponding decomposition and model assignments selected by the agent over time, while Figure \ref{fig:advdiff-adaptive-solerr} shows the resulting hybrid solution and pointwise error.  The reader can observe from these figures that the RL agent makes frequent changes to the selected DD, particularly during the early portion of the simulation; however, these changes do not translate into an improved accuracy-cost tradeoff relative to the fixed-DD strategy considered in Section \ref{sec:fixed_sd}.

\begin{figure}[ht!]
\centering
\includegraphics[width=\textwidth]{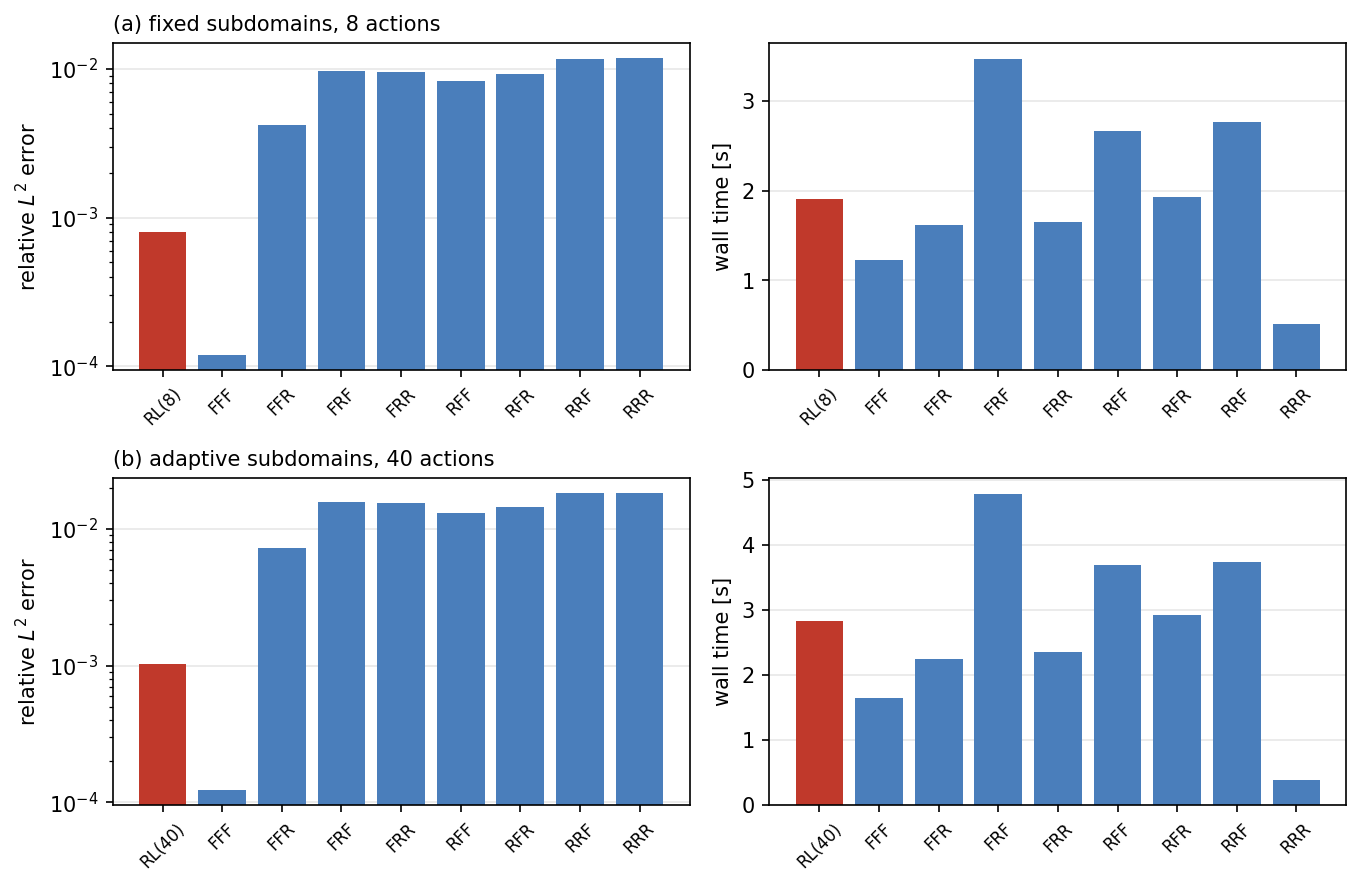}

\caption{Advection-diffusion benchmark, comparison of fixed and adaptive-DD. Relative $L^2$ error (left) and measured wall time (right) for the learned policy and the baseline FOM/ROM assignments. The top row corresponds to the fixed-DD with 8 actions, while the bottom row corresponds to the adaptive-DD with 40 actions. The learned policy is shown in red.}
\label{fig:advdiff-compare}
\end{figure}

Finally, Figure \ref{fig:advdiff-compare} provides a direct comparison of the accuracy and computational cost obtained with the fixed and adaptive-DD strategies, complementing the quantitative results in Tables \ref{tab:advdiff8} and \ref{tab:advdiff40}. In both cases, the learned policy achieves substantially lower error than the baseline FOM/ROM assignments containing at least one ROM. However, expanding the action space from 8 to 40 actions by allowing the agent to select the DD does not improve the learned policy. Compared with the fixed-DD policy, the adaptive-DD policy requires slightly fewer Schwarz iterations per decision window (2.73 vs. 3.00), but has a higher relative $L^2$ error ($1.03\times10^{-3}$ vs. $8.02\times10^{-4}$) and a substantially higher wall time (2.84 s vs. 1.91 s). We conclude that, for this benchmark, allowing the agent to select the DD in addition to the FOM/ROM assignment does not improve the overall accuracy-cost tradeoff. This result is favorable from an implementation standpoint, since dynamically changing the DD during a simulation is not feasible in current production implementations of O-SAM. The remainder of this work therefore focuses on online adaptation of the FOM/ROM assignment for a fixed-DD.  Nevertheless, this experiment demonstrates that the proposed RL framework can accommodate a larger action space in which both the model assignment and domain decomposition are selected online.




\subsection{Clamped linear elastic wave propagation benchmark}  \label{sec:clamped}
The second example considered herein is a linear elastic wave propagation benchmark implemented within the {\tt Norma.jl} open-source three-dimensional (3D) solid mechanics code \cite{norma}.  This test case is a variant of similar test cases considered in \cite{Mota:2022, barnett2022schwarzalternatingmethodseamless, tezaur2025hybrid}

Consider the equations of dynamic linear elasticity on a domain
$\Omega \subset \mathbb{R}^3$ over the time interval $t\in[0,T_{\text{final}}]$. 
Let $~u(~x,t) := \left( \begin{array}{ccc} u_x(~x,t), & u_y(~x,t), &u_z(~x,t) \end{array}\right) \in \mathbb{R}^3$ denote the displacement field. In the absence of body forces, the
governing equations are
\begin{equation}
\rho \ddot{~u}
-
\nabla\cdot\boldsymbol{\sigma}(~u)
=
\mathbf{0}
\qquad
\text{in } \Omega\times(0,T],
\label{eq:linear_elasticity}
\end{equation}
where $\rho$ is the material density and
$\boldsymbol{\sigma}$ is the Cauchy stress tensor. For a homogeneous,
isotropic, linear elastic material,
\begin{equation}
\boldsymbol{\sigma}(~u)
=
\lambda\,\mathrm{tr}\!\left(\boldsymbol{\varepsilon}(~u)\right)
~I
+
2\mu\,\boldsymbol{\varepsilon}(~u),
\label{eq:linear_elastic_stress}
\end{equation}
with infinitesimal strain tensor
\begin{equation}
\boldsymbol{\varepsilon}(~u)
=
\frac{1}{2}
\left(
\nabla~u
+
\nabla ~u^{T}
\right).
\label{eq:linear_strain}
\end{equation}
Here, $\lambda$ and $\mu$ are the Lam\'e parameters, which may be
expressed in terms of the Young's modulus $E$ and Poisson ratio $\nu$ as
\begin{equation}
\lambda
=
\frac{E\nu}{(1+\nu)(1-2\nu)},
\qquad
\mu
=
\frac{E}{2(1+\nu)}.
\label{eq:lame_parameters}
\end{equation}

For the clamped linear elastic wave propagation benchmark, the computational
domain is a slender 3D beam geometry, so that $\Omega = 
\left(-5\times10^{-4},5\times10^{-4}\right)
\times
\left(-5\times10^{-4},5\times10^{-4}\right)
\times
(-0.5,0.5)$ m.  We utilize the following material properties: 
$E = 1$ GPa, $\nu = 0$, and $\rho = 1000$ kg/m$^3$.
The problem is evolved over $0 \leq t \leq T_{\text{final}}$, with $T_{\text{final}} = 2 \times 10^{-3}$ s.
The spatial domain is discretized using eight-node hexahedral elements with an axial mesh spacing of $\Delta z=0.002$ m, 
and with $\Delta x = \Delta  y = 10^{-3} $, making the mesh exactly one element thick in the $x$- and $y$-directions.
All overlapping subdomain meshes employed in the O-SAM calculations are conformal. The resulting semi-discrete equations are integrated in time using the implicit Newmark-$\beta$ time integration scheme, with $\gamma=\frac{1}{2}$ and $\beta=\frac{1}{4}$, with a time step of $\Delta t=3.125\times10^{-6}$ s.

Herein, for simplicity, we wish to restrict the dynamics to the axial ($z$) direction in order to obtain a 1D problem solved using the 3D code, {\tt Norma.jl}.  
In order to accomplish this, homogeneous
Dirichlet boundary conditions are imposed on the transverse displacement
components:
\begin{equation}
    \begin{array}{cc}
        u_x(~x,t) = 0,
& \text{at } x=\pm 5\times10^{-4},  \\
      u_y(~x,t) = 0,
& \text{at } y=\pm 5\times10^{-4},
    \end{array}
\end{equation}
The beam is additionally clamped in the axial direction at its two ends,
giving
\begin{equation}
\begin{array}{cc}
u_z(~x,t) = 0, & \text{at } z=\pm0.5.
\end{array}
\end{equation}
The remaining unconstrained displacement components are subject to
homogeneous Neumann boundary conditions.

The initial conditions for our benchmark are: 
\begin{equation}
~u(~x,0)
=
\begin{pmatrix}
0\\
0\\
f(z)
\end{pmatrix},
\qquad
\dot{~u}(~x,0)
=
\begin{pmatrix}
0\\
0\\
f'(z)
\end{pmatrix}
\label{eq:initial_conditions}
\end{equation}
where $f(z) \in \mathbb{R}$ specifies the initial axial displacement profile, and $f'(z):=\frac{df}{dz}$.  Here, we consider initial axial displacement profiles of the form  
\begin{equation} \label{eq:clamped_ic}
    f(z) = a\exp\left( -\frac{z^2}{2s^2}\right) ,
\end{equation}
for $a, s \in \mathbb{R}$,
so that 
\begin{equation} \label{eq:clamped_ic_velo}
    f'(z) =  -\frac{ac}{s^2} z\exp\left(-\frac{z^2}{2s^2} \right),
\end{equation}
where $c = \sqrt{E/\rho}$ m/s is the speed of the traveling wave.

To assess the ability of the learned policy to generalize across different initial conditions, we consider a family of problems in which both the initial pulse location and width are varied. Specifically, we generalize the initial condition in \eqref{eq:clamped_ic} by defining
\begin{equation} \label{eq:clamped_ic_varied}
f(z)=a\exp\left(-\frac{(z-z_0)^2}{2s^2}\right).
\end{equation}
with the corresponding initial velocity obtained by differentiating $f(z)$ with respect to $z$, as in \eqref{eq:clamped_ic_velo}.  We consider seven equally spaced values of the pulse location $z_0\in[-0.05,0.10]$ m, with spacing $0.025$ m, and three pulse widths $s\in\{0.02,0.03,0.04\}$ m, yielding a total of 21 problem instances. The pulse amplitude is fixed at $a=1.0\times10^{-3}$ m.

For all configurations considered below, each subdomain-local OpInf ROM uses a reduced basis of dimension $r=10$. As in Section \ref{sec:advdiff}, the ROM dimension was selected by testing a range of candidate values, with $r=10$ being the smallest dimension that provided acceptable ROM accuracy. The OpInf least-squares problem \eqref{eq:opinf-learning} uses Tikhonov regularization with regularization parameter $\eta=10^{-8}$. The ROMs are constructed using FOM snapshot data from all 21 problem instances considered below, making the OpInf ROMs reproductive with respect to the problem instances considered in this benchmark. Snapshots are collected at fixed, uniformly spaced output times with a snapshot interval of $\Delta t_{\mathrm{snap}}=6.25\times10^{-6}$ s, corresponding to every two Newmark time steps. 

Based on the results of Section \ref{sec:adaptive_sd}, we restrict the present study to fixed-DDs. For the advection-diffusion benchmark (Section \ref{sec:advdiff}), allowing the RL agent to additionally select among multiple candidate domain decompositions did not improve the overall accuracy-cost tradeoff relative to adapting only the FOM/ROM assignment. Moreover, dynamically changing the DD during a simulation is not currently supported within 3D code implementations such as {\tt Norma.jl}. 

For the subdomain-local OpInf ROM training, each episode is divided uniformly into 32 decision windows, corresponding to 20 time steps per decision window. The state supplied to the DQN consists of the displacement field together with the action selected in the preceding decision window. For the reward function \eqref{eq:reward}, $\mathrm{cost}_k$ is the computational cost incurred during decision window $k$.  
Since the linear elastic wave propagation benchmark is a \textit{dynamics} problem, there are three solution fields of interest: the displacement, the velocity and the acceleration, denoted by $~u$, $~v$ and $~a$, respectively; this means that all three fields should go into the $\mathrm{err}_k$ term in \eqref{eq:reward}.
Toward this effect, to evaluate $\mathrm{err}_k$, the displacement, velocity, and acceleration solutions from the individual subdomains are first assembled into global fields $~u_k$, $~v_k$, and $~a_k$ on the global grid spanning the full domain $\Omega$, with the solutions from neighboring subdomains averaged at grid points in the overlap regions. The error is then defined as
\begin{equation} \label{eq:err_clamped}
     \mathbf{err}_k= \sqrt{ \frac{ \|~u-~u^{*}\|_2^2+ \|~v-~v^{*}\|_2^2+ \|~a-~a^{*}\|_2^2 }{ \|~u^{*}\|_2^2+ \|~v^{*}\|_2^2+ \|~a^{*}\|_2^2 }},
\end{equation}
where $(~u^{*},~v^{*},~a^{*})$ denotes the corresponding all-FOM O-SAM reference solution and $\|\cdot\|_2$ denotes the discrete $\ell_2$ norm. We set $\alpha=200$ and $\beta=1.5$ in \eqref{eq:reward}. As before, these parameters were fixed prior to training by sweeping each over several orders of magnitude and selecting values for which no single term (cost, accuracy, or switching) dominates the reward.

In what follows, two DDs are considered: (i) a two subdomain decomposition (Section \ref{sec:clamped_2sd}), and a (ii) three subdomain decomposition (Section \ref{sec:clamped_3sd}). For each decomposition, the RL agent selects the FOM/ROM assignment among the subdomains, yielding $2^2=4$ and $2^3=8$ possible actions, respectively.  
The high-fidelity reference solution introduced in Section \ref{sec:o-sam} is the corresponding all-FOM O-SAM solution and includes the displacement, velocity and acceleration fields, as in \eqref{eq:err_clamped}. Thus, the reference is obtained using the FF assignment for the two subdomain decomposition and the FFF assignment for the three subdomain decomposition. We adopt an all-FOM O-SAM reference because this problem is implemented in the production-oriented \({\tt Norma.jl}\) solid mechanics code, where, for problems of practical interest, construction of a single monolithic mesh may be difficult or impractical. 
  This reference is used to evaluate the error term \eqref{eq:err_clamped} in the reward during training and to assess the accuracy of the learned policy and fixed FOM/ROM assignments during predictive testing.

Separate DQN policies are trained for the two configurations.
Of the 21 problem instances, 11 are used for RL training and the remaining 10 are reserved for testing the learned policies. The split is constructed so that the test cases represent combinations of pulse location and width not encountered during RL training, but within the parameter range spanned by the training set. The two subdomain policy is trained for 1000 episodes, while the three subdomain policy is trained for 5000 episodes.
Thus, although the OpInf ROMs are reproductive for this benchmark, the RL policies are evaluated predictively on the held-out test cases.

\subsection{Two subdomain study} \label{sec:clamped_2sd}

We first consider a decomposition of the 3D beam into two overlapping subdomains, $\Omega_1$ and $\Omega_2$. Each subdomain spans the full cross-section of the beam, with axial extents $z\in[-0.50,-0.20]$ m for $\Omega_1$ and $z\in[-0.30,0.50]$ m for $\Omega_2$. Thus, the two subdomains overlap over $z\in[-0.30,-0.20]$ m. The corresponding RL action space consists of the four possible FOM/ROM assignments to the two subdomains.  

\begin{figure}[ht!]
\centering
\includegraphics[width=\textwidth]{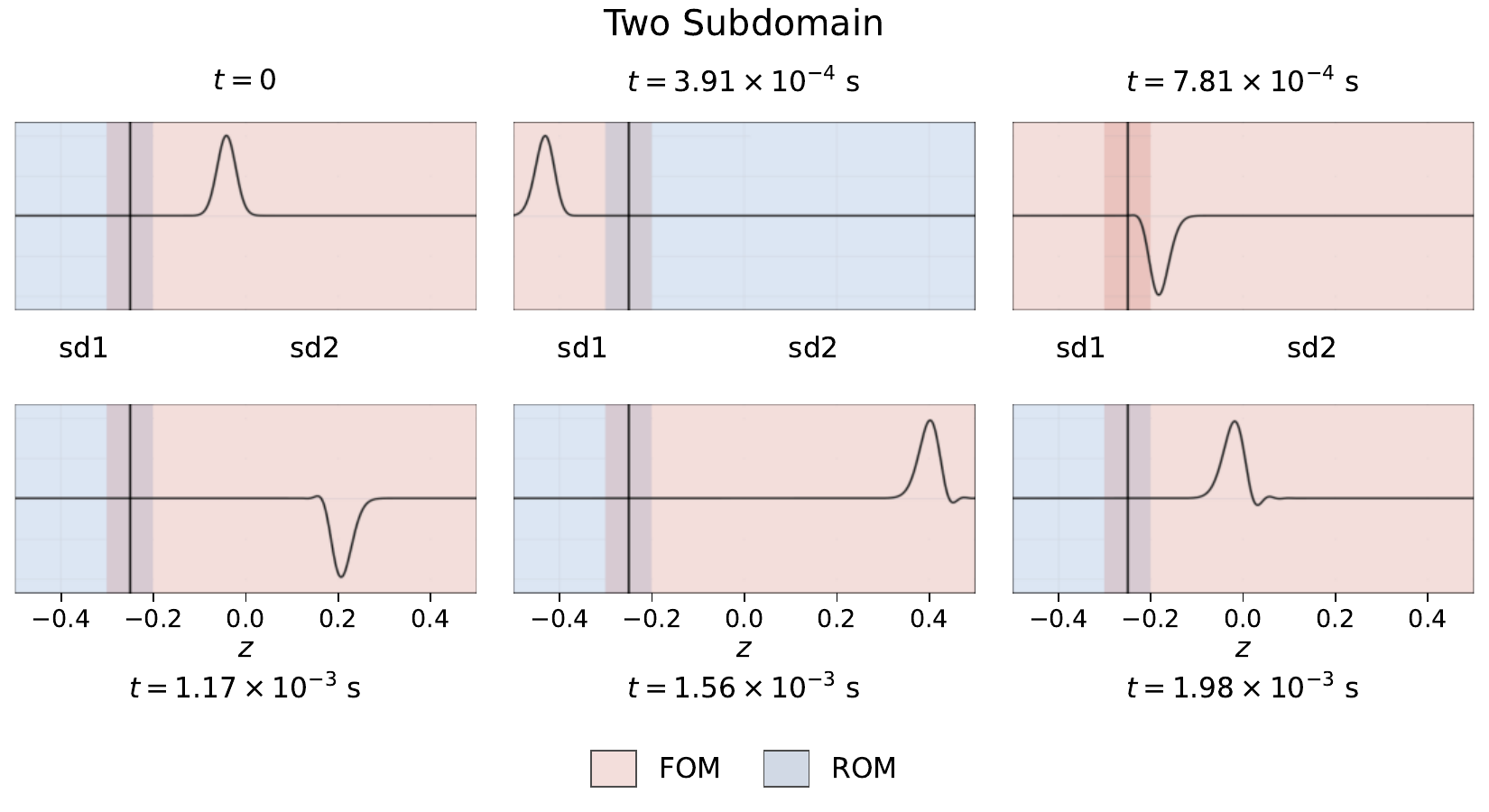}
\caption{Linear elastic wave benchmark, two subdomain case. Snapshots of the axial
displacement $u_z(z,t)$ at six times over the episode, for the predictive
test case $z_0=-0.0375$ m. Colors indicate the FOM/ROM assignment
selected by the trained DQN policy (red: full order; blue: reduced order),
and the vertical black line marks the divide between $\Omega_1$
($z\in[-0.50,-0.20]$ m) and $\Omega_2$ ($z\in[-0.30,0.50]$ m).}
\label{fig:wave-2sd}
\end{figure}

The DQN policy  for this configuration is trained for 1000 episodes using the training problem instances described above. We then evaluate the trained policy predictively on the held-out test cases. Figure \ref{fig:wave-2sd} shows the axial displacement $u_z(z,t)$ and the FOM/ROM assignments selected by the policy at six representative times for a held-out test case with $z_0=-0.0375$ m.  At $t=0$, the pulse is located primarily in $\Omega_2$, and the policy assigns the FOM to $\Omega_2$ and the ROM to $\Omega_1$. As the pulse propagates to the left and enters $\Omega_1$, the policy assigns the FOM to $\Omega_1$ and the ROM to $\Omega_2$. Following reflection from the clamped boundary at $z=-0.5$ m, the pulse propagates back toward $\Omega_2$. As the reflected wave moves from $\Omega_1$ toward $\Omega_2$, the policy temporarily assigns the FOM to both subdomains. Once the wave has propagated farther into $\Omega_2$, the policy returns to a ROM-FOM assignment. The policy therefore adapts the FOM/ROM assignment in response to the propagation of the localized wave between the two subdomains.  This behavior is expected: the FOM is preferentially assigned to the subdomain containing the propagating wave, while the ROM is used in regions where the solution is comparatively smooth.

\begin{table}[ht!]
\centering
\begin{tabular}{lcc}
\hline
policy & relative $L^2$ error & wall time [s] \\
\hline
RL (4 actions) & \textbf{$1.59\times 10^{-2}$} & \textbf{$49.77$} \\
FF                      & $0$                           & $52.91$ \\
RF                      & $2.43\times 10^{-1}$          & $45.00$ \\
FR                      & $2.67\times 10^{-1}$          & $42.09$ \\
RR                      & $2.69\times 10^{-1}$          & $33.23$ \\
\hline
\end{tabular}
\caption{Linear elastic wave benchmark, two subdomain case. Results for the learned policy and the four possible fixed FOM/ROM assignments. ‘F’ and ‘R’ denote full order and reduced order subdomain models, respectively, ordered from left to right. The relative $L^2$ error is computed with respect to
the FF assignment, which therefore has zero error by construction. The results correspond to the predictive test case $z_0=-0.0375$ m shown in Figure \ref{fig:wave-2sd}.}
\label{tab:wave-2sd-wall}
\end{table}

Table~\ref{tab:wave-2sd-wall} compares the learned policy with each of the four fixed FOM/ROM assignments. The learned policy achieves a relative $L^2$ error of $1.59\times10^{-2}$, more than an order of magnitude smaller than that of the most accurate fixed assignment containing at least one ROM (RF), whose error is $2.43\times10^{-1}$. The learned policy also reduces the wall time from 52.91 s for the all-FOM (FF) assignment to 49.77 s. 
The all-FOM relative error is identically zero, as the all-FOM solution is utilized as the reference solution in the $L^2$ error calculation.  
As expected, the all-ROM (RR) assignment has the lowest wall time, 33.23 s, but its relative $L^2$ error of $2.69\times10^{-1}$ is approximately 17 times larger than that of the learned policy. Thus, as in the advection-diffusion benchmark (Section \ref{sec:advdiff}), the learned policy provides a compromise between the high accuracy of the all-FOM assignment and the lower computational cost of assignments involving fixed ROMs.

\subsection{Three subdomain study} \label{sec:clamped_3sd}

We next consider a decomposition of the 3D beam into three overlapping subdomains, $\Omega_1$, $\Omega_2$, and $\Omega_3$. Each subdomain spans the full cross-section of the beam, with axial extents $z\in[-0.50,-0.10]$ m, $z\in[-0.20,0.20]$ m, and $z\in[0.10,0.50]$ m, respectively. Thus, adjacent subdomains overlap over $z\in[-0.20,-0.10]$ m and $z\in[0.10,0.20]$ m. The corresponding RL action space consists of the $2^3=8$ possible FOM/ROM assignments to the three subdomains.

\begin{figure}[ht!]
\centering
\includegraphics[width=\textwidth]{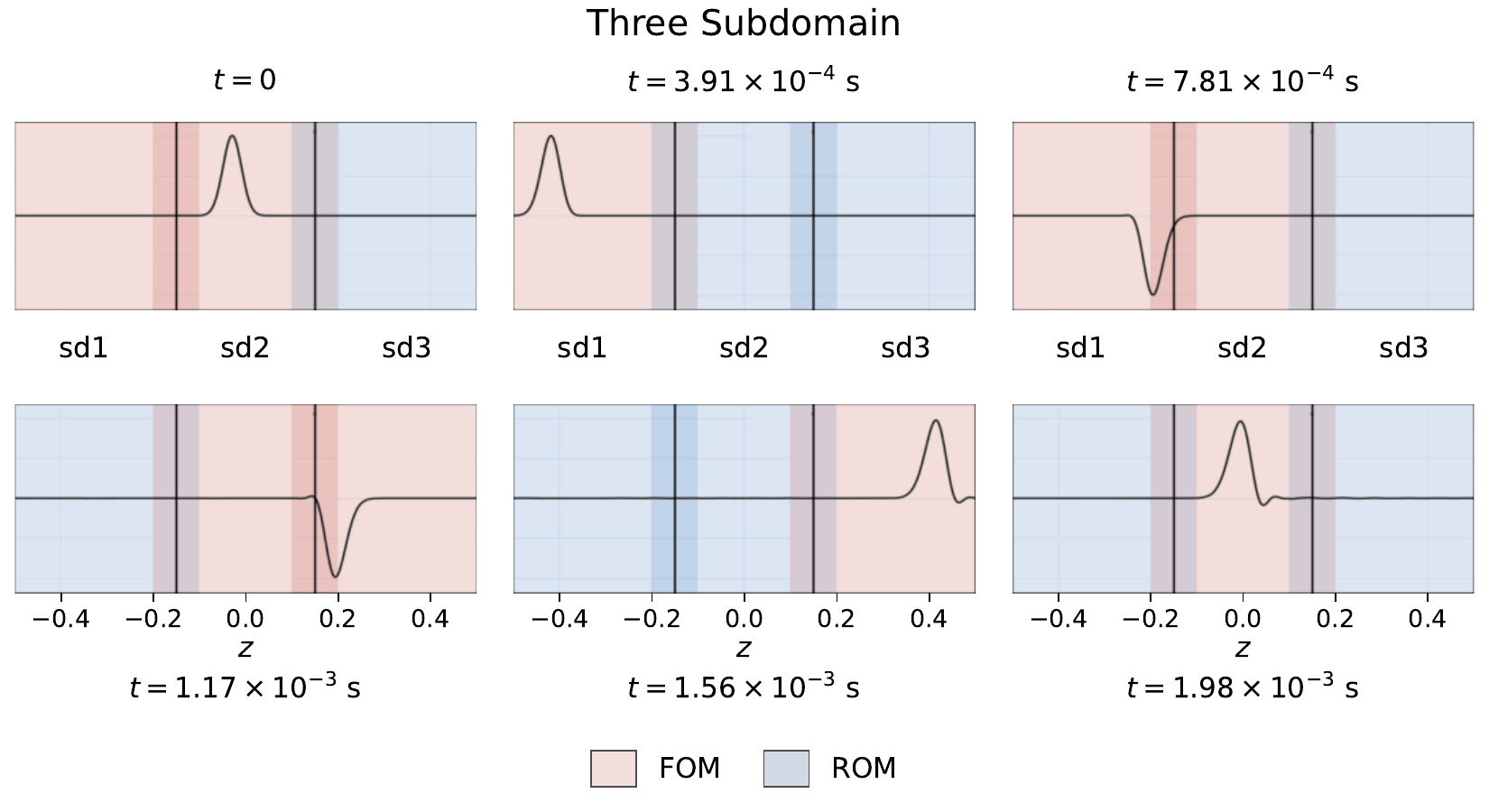}
\caption{Linear elastic wave benchmark, three subdomain case. Snapshots of the axial
displacement $u_z(z,t)$ at the same six times as Figure~\ref{fig:wave-2sd},
for the predictive test case $z_0=-0.025$ m, $s=0.02$ m. Shading and
vertical lines are as in Figure~\ref{fig:wave-2sd}, now marking the two
divides between $\Omega_1$ ($z\in[-0.50,-0.10]$ m), $\Omega_2$
($z\in[-0.20,0.20]$ m), and $\Omega_3$ ($z\in[0.10,0.50]$ m).}
\label{fig:wave-3sd}
\end{figure}

The DQN policy for the three subdomain configuration is trained for 5000 episodes using the training problem instances described above. As for the two subdomain case, we evaluate the trained policy predictively on the held-out test cases. Figure \ref{fig:wave-3sd} shows the axial displacement $u_z(z,t)$ and the FOM/ROM assignments selected by the policy at six representative times for a held-out test case with $z_0=-0.025$ m and $s=0.02$ m.  At $t=0$, the pulse is located primarily in $\Omega_2$, and the policy assigns the FOM to $\Omega_2$ and ROMs to $\Omega_1$ and $\Omega_3$. As the pulse propagates toward the left end of the beam, the policy assigns the FOM to $\Omega_1$ and ROMs to the other two subdomains. Following reflection from the left clamped boundary, the wave propagates toward the right end of the beam, with the FOM assignment shifting from $\Omega_1$ and $\Omega_2$, to $\Omega_2$ and $\Omega_3$, and finally to $\Omega_3$ as the wave approaches the right boundary. After reflection, the FOM is again assigned to $\Omega_2$ as the wave propagates back toward the center.
As in the two subdomain case (Section \ref{sec:clamped_2sd}), the results described above exhibit the expected behavior: the policy preferentially assigns the FOM to the subdomain or subdomains containing the propagating wave and uses the ROM elsewhere. The three subdomain example demonstrates that this behavior is retained as the number of subdomains and possible FOM/ROM assignments increases.

\begin{table}[ht!]
\centering
\begin{tabular}{lcc}
\hline
policy & relative $L^2$ error & wall time [s] \\
\hline
RL (8 actions) & \textbf{$3.90\times 10^{-3}$} & \textbf{$54.04$} \\
FFF                     & $0$                           & $62.63$ \\
FFR                     & $1.54\times 10^{-1}$          & $60.03$ \\
FRF                     & $2.23\times 10^{-1}$          & $67.53$ \\
RRF                     & $2.58\times 10^{-1}$          & $42.42$ \\
FRR                     & $2.58\times 10^{-1}$          & $55.72$ \\
RRR                     & $2.94\times 10^{-1}$          & $42.10$ \\
RFF                     & $4.05\times 10^{-1}$          & $61.39$ \\
RFR                     & $8.48\times 10^{-1}$          & $50.22$ \\
\hline
\end{tabular}
\caption{Linear elastic wave benchmark, three subdomain case. Results for the learned policy and the eight possible fixed FOM/ROM assignments. ‘F’ and ‘R’ denote full order and reduced order subdomain models, respectively, ordered from left to right. The relative $L^2$ error is computed with
respect to the FFF assignment, which therefore has zero error by
construction. The results correspond to the predictive test case $z_0=-0.025$ m and $s=0.02$ m shown in Figure \ref{fig:wave-3sd}.}
\label{tab:wave-3sd-wall}
\end{table}

Table~\ref{tab:wave-3sd-wall} compares the learned policy with each of the eight fixed FOM/ROM assignments. The learned policy achieves a relative $L^2$ error of $3.90\times10^{-3}$, approximately 40 times smaller than that of the most accurate fixed assignment containing at least one ROM (FFR), whose error is $1.54\times10^{-1}$. The learned policy also reduces the wall time from 62.63 s for the all-FOM (FFF) assignment to 54.04 s. 
As for the two subdomain version of this problem (Section \ref{sec:clamped_2sd}), the all-FOM relative error is identically zero, due to the fact that the all-FOM solution is utilized as the reference solution  in the $L^2$ error calculation.  
As expected, the all-ROM (RRR) assignment has the lowest wall time, 42.10 s, but its relative $L^2$ error of $2.94\times10^{-1}$ is approximately 75 times larger than that of the learned policy. Thus, the three subdomain results exhibit the same overall behavior as the two subdomain results: the learned policy retains accuracy much closer to that of the all-FOM solution while reducing its computational cost.

\section{Conclusions} \label{sec:conc}

In this work, we have developed an RL-based approach for online adaptation of hybrid FOM-ROM models coupled through the overlapping Schwarz alternating method. The objective is to address a limitation of conventional hybrid Schwarz coupling, in which the model assigned to each subdomain is typically selected \textit{a priori} and remains fixed throughout the simulation. Such a fixed assignment may be inadequate for transient problems involving moving or evolving features, since the regions requiring high-fidelity resolution can change over time. To address this challenge, we employed DQNs to learn policies that dynamically select between subdomain-local FOMs and pre-trained OpInf ROMs as the solution evolves. The policies are trained offline using a reward function that balances solution accuracy and computational cost while penalizing unnecessary model switching, and are subsequently deployed online without requiring access to a reference FOM solution.

We studied our proposed approach using two transient benchmark problems.  First, a 1D advection-diffusion problem with a moving front was used to study online FOM/ROM switching for both fixed and adaptive domain decompositions and to examine the effect of the switching penalty in the reward function. For this first benchmark, both the OpInf ROMs and the learned RL policy were evaluated predictively.  Second, a 3D linear elastic wave propagation problem implemented in the {\tt Norma.jl} code \cite{norma} was used to evaluate the approach on a solid mechanics test case with two  and three subdomain decompositions, in which the initial pulse location and width are varied and the learned policies are evaluated predictively on unseen problem instances. For this second benchmark, the OpInf ROMs were reproductive, while the RL policies were evaluated predictively on problem instances withheld from RL training.

For the advection-diffusion benchmark, the learned policy successfully adapted the FOM/ROM assignment as the moving front propagated through the domain. For the fixed domain decomposition, it achieved a relative $L^2$ error of $8.02\times10^{-4}$, approximately five times smaller than that of the most accurate fixed FOM/ROM assignment containing at least one ROM, while also reducing the wall time relative to the all-FOM coupled solution. The switching penalty was found to play an important role in regularizing the learned policy, reducing unnecessary FOM/ROM switching without changing the underlying objective of balancing accuracy and computational cost. Allowing the agent to adapt the DD in addition to the FOM/ROM assignment did not improve the overall accuracy-cost tradeoff: although the adaptive-DD policy required slightly fewer Schwarz iterations, it exhibited both higher error and higher wall time than the learned policy for the fixed domain decomposition. These results suggest that online adaptation of the FOM/ROM assignment within a fixed-DD setting provides the more effective and practically realizable strategy for the problems considered here.

The 3D linear elastic wave propagation benchmark demonstrated that the same approach can be applied within the {\tt Norma.jl} solid mechanics code \cite{norma}. For both the two and three subdomain configurations, the learned policies exhibited the expected behavior, preferentially assigning the FOM to subdomains containing the propagating wave and using ROMs elsewhere. As the wave propagated through the beam and reflected from the clamped boundaries, the policies adapted the FOM/ROM assignments accordingly. The improvement relative to fixed FOM/ROM assignments was particularly pronounced for this benchmark, with the learned policies achieving errors more than an order of magnitude smaller than the most accurate fixed assignments containing at least one ROM, while also reducing the wall time relative to the all-FOM assignments.  Importantly, these results were obtained for held-out combinations of pulse location and width, indicating that the learned policies can generalize their model selection behavior to initial conditions not encountered during the training of the RL agent.

The computational savings relative to the all-FOM coupling observed for both benchmarks are moderate. We attribute this to two features of the test problems considered. First, the FOMs are small, so the cost reduction obtained by replacing a FOM with a ROM in a given subdomain is limited, and overheads that do not shrink with the ROM, such as Schwarz iterations, interface data transfer and switching between models, make up a relatively large share of the total cost. For larger FOMs, the relative cost of these overheads decreases and the ROM-to-FOM speedup in each subdomain increases. Second, because each problem has a propagating front, at least one subdomain must use the FOM at any given time, so the overall savings are bounded by the fraction of the domain in which a ROM can be used. We therefore expect the proposed approach to be most beneficial for large-scale problems in which the features requiring high-fidelity resolution occupy a small, moving portion of the domain. Realizing these savings in parallel computations will additionally require load balancing, since FOM and ROM subdomains have very different computational costs, and the number of processes assigned to each subdomain may need to change as the model assignment changes.

Several directions for future work follow naturally from the present study. First, the implementation developed for the linear elastic wave propagation benchmark provides a workflow for applying the proposed adaptive FOM/ROM coupling strategy to fully 3D problems in {\tt Norma.jl}. A natural next step is therefore to apply this capability to more complex problems without a clear propagating front, in which the regions requiring high-fidelity resolution are not known in advance and may evolve in less predictable ways, including nonlinear and truly 3D solid mechanics problems involving richer spatial and temporal dynamics. For the benchmarks considered herein, the propagating front made the expected behavior of the policy clear, which allowed us to verify that the learned policy is physically intuitive; we expect the main value of the RL approach to lie in problems for which no such intuitive strategy is available. Second, the OpInf ROMs considered in the present work are constructed offline and remain fixed throughout deployment. An important direction for future work is to enable these ROMs to adapt online using newly available FOM information generated during the hybrid simulation. Online adaptation of data-driven ROMs remains an open research area; one promising approach to investigate is the streaming OpInf methodology of Koike et al. \cite{koike2026streamingoperatorinferencemodel}, which enables reduced operators to be updated incrementally as new data become available.  Finally, the present work employs DQN to learn the adaptive model-selection policy. Future studies could investigate more advanced reinforcement learning approaches, including methods designed for larger or more structured action spaces, as the number of subdomains and available modeling choices increases. The formulation of the reward could also be refined. For example, normalizing $\mathrm{cost}_k$ in \eqref{eq:reward} by the cost of the all-FOM configuration would make all terms in the reward dimensionless and the choice of $\alpha$ independent of the hardware used, while a reward based on a user-specified error tolerance is a natural alternative. In addition, since all baselines considered herein use fixed FOM/ROM assignments, it would be useful to compare the learned policy against a randomized baseline in which each subdomain is assigned the FOM with some probability $p$ at each decision window, in order to separate the benefit of the learned policy from that of adaptation alone. The action space could also be enriched with ROMs of several dimensions in each subdomain, so that the agent can choose among multiple fidelity levels instead of a single FOM/ROM pair, which could further reduce the computational cost. 

\section*{Acknowledgements} \label{sec:acknowl}

Support for this work was received through Sandia National Laboratories' Laboratory Directed Research and Development (LDRD) program and through the U.S. Department of Energy, Office of Science, Office of Advanced Scientific Computing Research, Mathematical Multifaceted Integrated Capability Centers (MMICCs) program, under Field Work Proposal 22025291 and the Multifaceted Mathematics for Predictive Digital Twins (M2dt) project. Additionally, the writing of this manuscript was funded in part by Irina Tezaur’s Presidential Early Career Award for Scientists and Engineers (PECASE).

The authors wish to thank Chris Wentland for valuable discussions and insightful advice that helped shape the research presented in this paper.  The authors additionally wish to thank Alejandro Mota for his assistance in using and developing the {\tt Norma.jl} code.

Sandia National Laboratories is a multi-mission laboratory managed and operated by National Technology and Engineering Solutions of Sandia, LLC., a wholly owned subsidiary of Honeywell International, Inc., for the U.S. Department of Energy’s National Nuclear Security Administration under contract DE-NA0003525.

\bibliographystyle{siam}
\bibliography{TrishitMondal}

\end{document}